\PassOptionsToPackage{table}{xcolor}
\documentclass[numbers]{fairmeta}

\usepackage{multirow}
\usepackage{graphicx}
\usepackage{array}
\usepackage{amsmath,amssymb,amsfonts}
\usepackage{subcaption}
\usepackage{booktabs}
\usepackage{enumitem}
\usepackage{fontawesome5}
\usepackage{algorithm}
\usepackage{algorithmic}
\usepackage{makecell}
\usepackage{colortbl}
\usepackage{xspace}
\usepackage{pifont}
\usepackage{wrapfig}

\def\appref#1{Appendix~\ref{#1}}
\def\apprefs#1#2{Appendices~\ref{#1}--\ref{#2}}

\definecolor{maePurple}{HTML}{6B3FA0}
\definecolor{refRed}{HTML}{C62828}
\definecolor{citationBlue}{HTML}{6FA8DC}
\definecolor{promptBlue}{HTML}{EAF3FF}
\definecolor{promptMint}{HTML}{EAF7F2}
\definecolor{promptGold}{HTML}{FFF6E6}
\definecolor{maeshade}{HTML}{F3EDF9}

\hypersetup{
  colorlinks=true,
  linkcolor=refRed,
  citecolor=citationBlue,
  urlcolor=citationBlue
}

\newcommand{\namett}[1]{{\fontfamily{lmtt}\selectfont #1}}
\newcommand{\purplettbf}[1]{{\fontfamily{lmtt}\selectfont\textbf{\textcolor{maePurple}{#1}}}}
\newcommand{\mathnamett}[1]{\text{\normalfont\fontfamily{lmtt}\selectfont #1}}
\newcommand{\mathpurplettbf}[1]{\textcolor{maePurple}{\text{\normalfont\fontfamily{lmtt}\selectfont\bfseries #1}}}
\newcommand{\ourmethod}{\purplettbf{MAE}\xspace}

\newcommand{\maed}{\purplettbf{MAE-D}\xspace}
\newcommand{\maec}{\purplettbf{MAE-C}\xspace}
\newcommand{\maemath}{\mathop{\mathpurplettbf{MAE}}\nolimits}
\newcommand{\maedmath}{\mathpurplettbf{MAE-D}}
\newcommand{\maecmath}{\mathpurplettbf{MAE-C}}
\newcommand{\topk}{\mathop{\mathnamett{TopM}}\nolimits}
\newcommand{\topkname}[1]{\namett{Top-#1}}
\newcommand{\modelname}[1]{\namett{#1}}
\newcommand{\benchname}[1]{\namett{#1}}
\newcommand{\subsetname}[1]{\namett{#1}}
\newcommand{\metricname}[1]{\namett{#1}}
\newcommand{\rqtag}[1]{\textbf{RQ#1}}

\newcounter{takeaway}
\newcommand{\takeaway}[2]{\refstepcounter{takeaway}\noindent\textbf{Takeaway \ding{#1}: #2}}
\newcommand{\llamaicon}{\raisebox{-0.18em}{\includegraphics[height=1.05em]{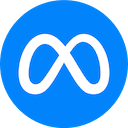}}}
\newcommand{\qwenicon}{\raisebox{-0.18em}{\includegraphics[height=1.05em]{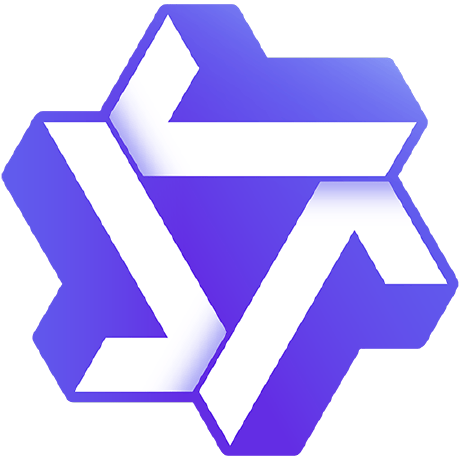}}}
\newcommand{\modelcell}[2]{\makecell[c]{#1\hspace{0.25em}{\footnotesize\textbf{#2}}}}
\newcommand{\mainmodelcell}[2]{\makecell[c]{#1\\[-0.2em]{\footnotesize\textbf{#2}}}}

\newcommand{\suiteheader}[3]{\makecell[c]{#1~\textbf{#2}\\[-0.15em]\scriptsize\subsetname{#3}}}
\newcommand{\mainmetricheader}[2]{\makecell[c]{\footnotesize\metricname{#1}\,{\scriptsize$#2$}}}
\newcommand{\shadebf}[1]{\cellcolor{maePurple!6}\textbf{#1}}
\newcommand{\baselinegroup}[1]{& \multicolumn{13}{l}{\rule{0pt}{1.25em}\textbf{\footnotesize #1}} \\[0.05em]}
\newcommand{\maeboxrule}{\arrayrulecolor{maePurple}\cmidrule(lr){2-14}\arrayrulecolor{black}}
\newcommand{\maemethodcell}[1]{\cellcolor{maeshade}#1}
\newcommand{\mainmaemethod}[2]{\maemethodcell{\makecell[l]{#1\\[-0.18em](\topkname{#2})}}}
\newcommand{\relgain}[1]{\textcolor{maePurple!75!black}{\scriptsize #1}}
\newcommand{\metricgain}[2]{\makecell[c]{\textbf{#1}\\[-0.18em]\relgain{#2}}}
\newcommand{\metricgainunder}[2]{\makecell[c]{\underline{#1}\\[-0.18em]\relgain{#2}}}
\newcommand{\metricgainplain}[2]{\makecell[c]{#1\\[-0.18em]\relgain{#2}}}

\newcommand{\draftrowmark}{\textsuperscript{\(\dagger\)}}

\newcommand{\promptcode}[1]{{\ttfamily\scriptsize #1}}
\newcommand{\goalicon}{\raisebox{-0.18em}{\includegraphics[height=1.05em]{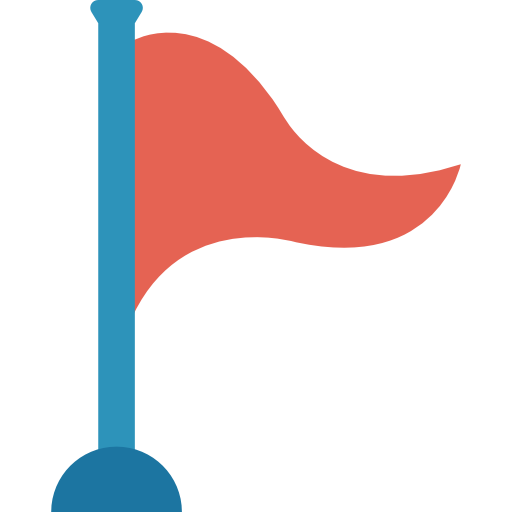}}}
\newcommand{\cubeicon}{\raisebox{-0.18em}{\includegraphics[height=1.05em]{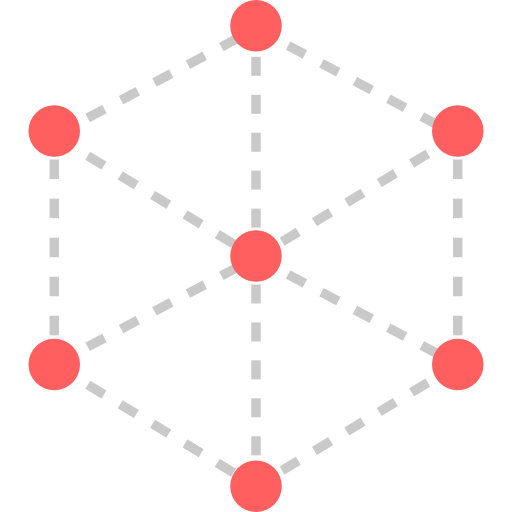}}}
\newcommand{\routeicon}{\raisebox{-0.18em}{\includegraphics[height=1.05em]{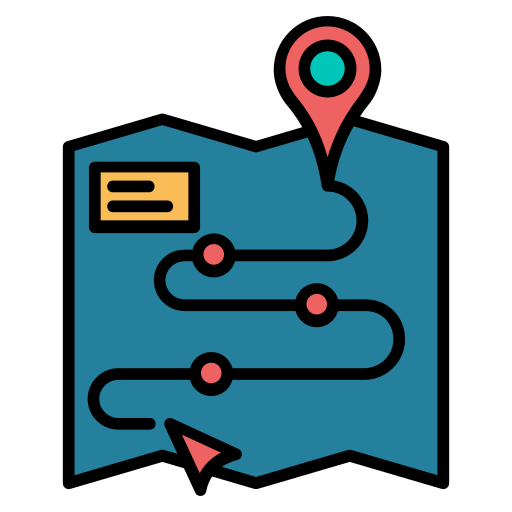}}}
\newcommand{\compicon}{\raisebox{-0.18em}{\includegraphics[height=1.05em]{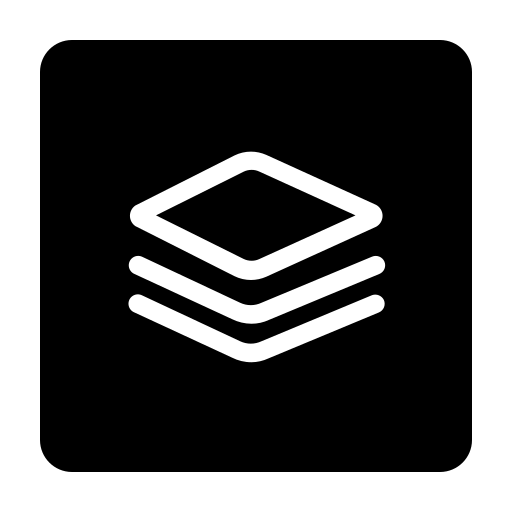}}}
\newcommand{\openvla}{\modelname{OpenVLA}\xspace}
\newcommand{\openvlaoft}{\modelname{OpenVLA-OFT}\xspace}
\newcommand{\qwenpiflow}{\modelname{QwenPI-Flow}\xspace}

\title{MAE I Trust Myself? Self-Evaluating VLA Action Generation with Markov Attention Entropy}

\author[2,3]{Aniri}
\author[5]{Chen Yilin}
\author[1,2,3]{Jinhe Bi}
\author[2]{Zengjie Jin}
\author[2,3]{Yujun Wang}
\author[4]{Yijun Tian}
\author[2,3]{Volker Tresp}
\author[1]{Fei Shen}
\author[1]{Tat-Seng Chua}
\author[2,3]{Yunpu Ma\,\faEnvelope}

\affiliation[1]{National University of Singapore}
\affiliation[2]{Ludwig Maximilian University of Munich}
\affiliation[3]{Munich Center for Machine Learning}
\affiliation[4]{Amazon}
\affiliation[5]{East China University of Science and Technology}

\vspace{0.5em}

\abstract{
  \begingroup
  \renewcommand{\ourmethod}{MAE\xspace}
  Vision-Language-Action models (VLAs) integrate visual perception, language instruction, and action generation into end-to-end policies across heterogeneous architectures. However, enabling VLAs to self-evaluate their action generation reliability without external supervision remains a major challenge. Existing methods either rely on expert annotations or estimate uncertainty only from output statistics, largely ignoring internal signals. In this work, we observe that internal visual modality entropy exhibits consistent distinctions between successful and failed tasks across heterogeneous VLAs. Although VLAs' architectures differ in their action generation, we show that they share a common latent action generation abstraction evolving under visual perception, language instruction, and State Input, which we formulate as a Conditional Generative Markov Chain. Based on this formulation, we propose \ourmethod (Markov Attention Entropy), a self-evaluation framework that directly converts internal attention signals into architecture-aware reliability scores, and introduce \benchname{LIBERO-Reflect}, a 4,000-episode benchmark combining 2,000 standard episodes and 2,000 challenging episodes across four subsets. Extensive experiments across heterogeneous VLA architectures and diverse scenarios show that \ourmethod consistently outperforms state-of-the-art baselines on AUPR, AUROC, and FPR@95. Code is available in \href{https://github.com/aniri15/MAE-Self-Evaluating-VLA.git}{this repository}.
  \endgroup
}
\date{\today}
\correspondence{\email{bijinhe@outlook.com}, \email{cognitive.yunpu@gmail.com}}

\begin{document}
\thispagestyle{firstheader}
\maketitle

\section{Introduction}
\begin{wrapfigure}[18]{r}{0.6\textwidth}
    \vspace{-1.8\baselineskip}
    \centering
    \includegraphics[width=\linewidth]{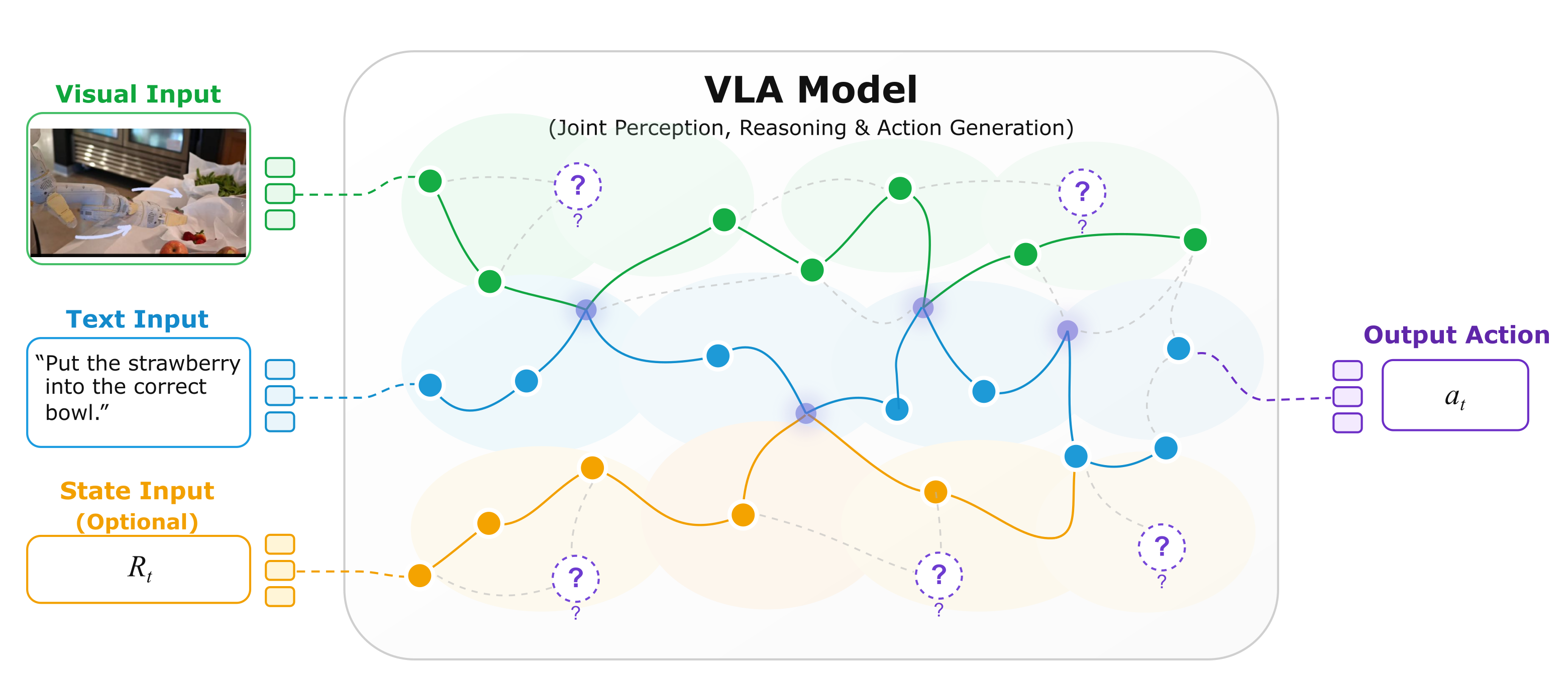}
    \caption{From multimodal conditioning to action generation in VLAs.
We view VLA action generation as a conditional generative Markov chain: visual observations, language instructions, and State Input jointly condition a sequence of latent action states that evolves toward final action.
Attention exposes how these latent states route information across modalities, and its entropy serves as a white-box signal for self-evaluating action-generation reliability.}
    \label{fig:vla-action-generation-overview}
\end{wrapfigure}

Vision-Language-Action models (VLAs) \citep{sapkota2026visionlanguageactionvlamodelsconcepts,kim2024openvlaopensourcevisionlanguageactionmodel,kim2025finetuningvisionlanguageactionmodelsoptimizing} translate visual observations and language instructions into executable robot actions by adapting pretrained Vision-Language Models to embodied control as illustrated in Figure~\ref{fig:vla-action-generation-overview}.
Their action-generation mechanisms, however, are heterogeneous.
In this work, we distinguish two broad action-generation families: Latent-Readout VLAs and Latent-Refinement VLAs.
The former evolves latent action representations and maps the final latent state to executable actions through a readout head; \modelname{OpenVLA} uses autoregressive discrete readout \citep{kim2024openvlaopensourcevisionlanguageactionmodel}, while \modelname{OpenVLA-OFT} uses continuous readout \citep{kim2025finetuningvisionlanguageactionmodelsoptimizing}.
The latter maintains an evolving action or action-trajectory representation and refines it iteratively under learned flow- or diffusion-style dynamics, as in PI-family policies \citep{intelligence2025pi05visionlanguageactionmodelopenworld}.
As VLAs generalize across robotic tasks \citep{liu2023liberobenchmarkingknowledgetransfer}, they also require reliable estimates of when their generated actions are likely to fail \citep{zhou2025liberoprorobustfairevaluation}.

\begin{wrapfigure}[30]{r}{0.5\textwidth}
  \vspace{-0.4\baselineskip}
  \centering
  \includegraphics[width=\linewidth]{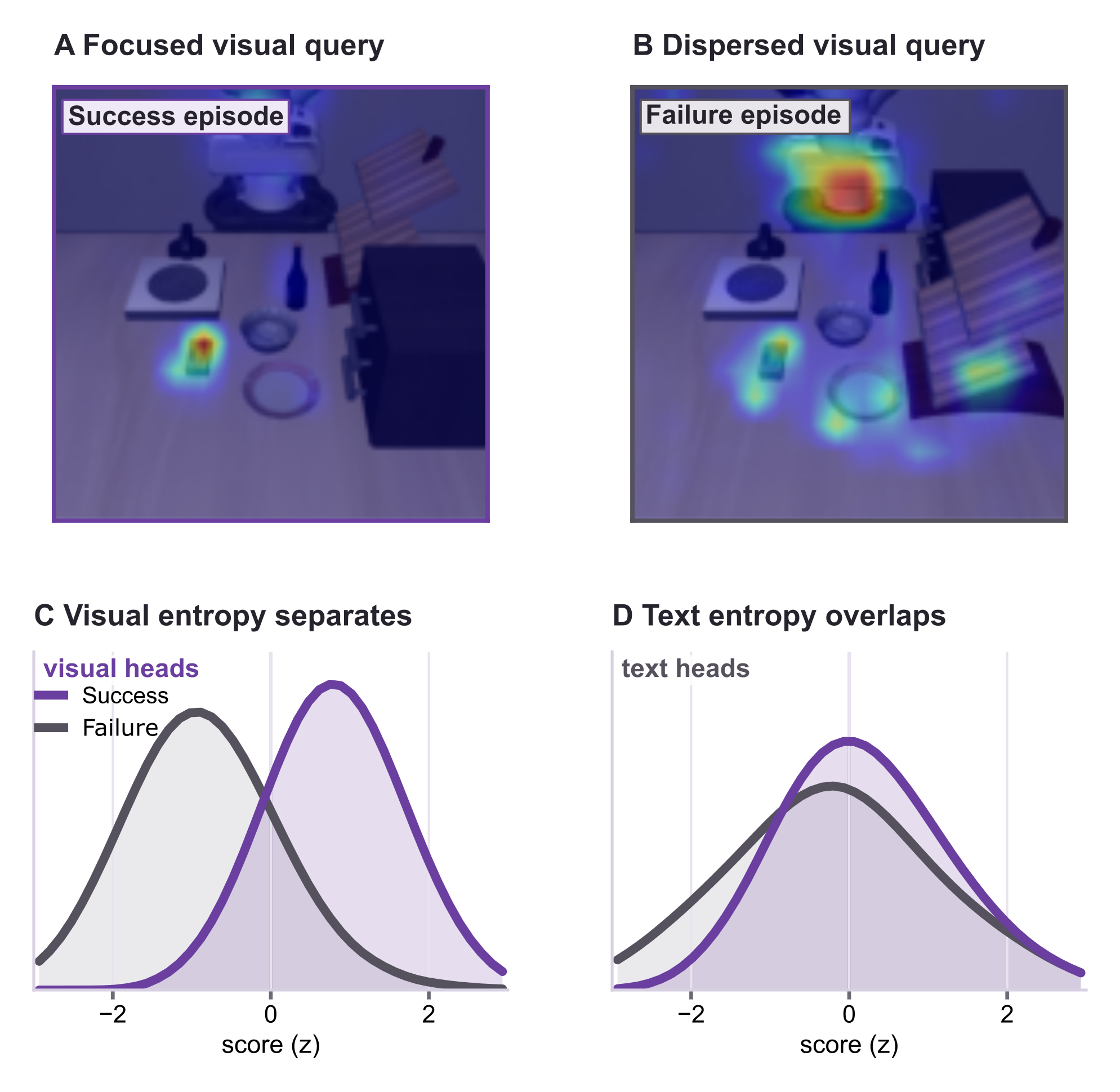}
  \caption{Attention entropy as an internal self-evaluation signal.
  Panels A--B show a successful visual query concentrated on the task-relevant object and a failed visual query dispersed across distractor regions.
  Panels C--D compare oriented episode-level score distributions from visual and text attention entropy on the same \subsetname{Reflect-Goal} episodes.
  Visual entropy yields a much clearer separation between successful and failed episodes than text entropy, supporting visual attention entropy as the action-relevant self-evaluation signal.}
  \label{fig:entropy-signal}
\end{wrapfigure}
The first line of reliability work relies on \textbf{External Supervision}.
Early studies introduced output-level uncertainty and quality metrics, often calibrated by expert annotations \citep{valle2025evaluatinguncertaintyqualityvisual}.
Later work moved beyond passive evaluation by using auxiliary models to detect, explain, verify, and recover from failures \citep{duan2024ahavisionlanguagemodeldetectingreasoning,qi2026selfrefiningvisionlanguagemodel,lin2025failsafereasoningrecoveryfailures,kwok2025robomonkeyscalingtesttimesampling,dai2025roverrobotrewardmodel,liu2025evovlaselfevolvingvisionlanguageactionmodel,chen2026reconvlauncertaintyguidedfailureawarevisionlanguageaction}.
These methods can improve VLA reliability, but their evaluation signal is supplied outside the policy.
This can increase deployment cost, introduce sensitivity to evaluator or annotation shift, and give limited access to the internal state transitions that produced a failed action \citep{casper2023openproblemsfundamentallimitations}.

A second line studies \textbf{Internal Self-evaluation}, where reliability is inferred from the VLA's own signals.
Existing internal methods can be grouped into black-box and white-box approaches.
Black-box methods rely on output statistics, such as multi-sample disagreement or self-uncertainty scores, to select actions without external supervision \citep{Bi2025CoTKineticsAT,Kadavath2022LanguageM,Wang2022SelfConsistencyIC}.
However, they can be over-confident and provide limited interpretability \citep{Bi2025CoTKineticsAT}.
White-box self-evaluation has been studied more systematically in Large Language Models (LLMs), where internal signals such as softmax confidence, temperature scaling, or activation patterns are used to estimate generation reliability without external supervision \citep{Bi2025CoTKineticsAT,si2023prompting,huang2023look,malinin2021uncertainty}.
For VLAs, related evidence remains limited and fragmented.
Existing studies show that internal signals can reflect path deviation, execution horizon, pathway specialization, or latent action bottlenecks~\citep{jeong2026visionlanguageactionmodelattentionheads, wang2026vlaknowslimits, grant2026featurescreatedequalmechanistic, haeon2025mechanisticinterpretabilitysteeringvisionlanguageaction, buurmeijer2026observingcontrollingfeaturesvisionlanguageaction, lian2026langforcebayesiandecompositionvision}.
What is still missing is a unified way to convert such internal signals into self-evaluation metrics that remain meaningful across heterogeneous VLA action-generation architectures.
This gap motivates the following question:

\begin{tcolorbox}[
  enhanced,
  sidebyside,
  sidebyside align=center,
  colframe=maePurple!75!black,
  colback=maeshade,
  boxrule=0.7pt,
  arc=2mm,
  left=1.5mm,
  right=2mm,
  top=1.5mm,
  bottom=1.5mm,
  lefthand width=0.085\linewidth,
  sidebyside gap=3mm,
  segmentation style={draw=maePurple!35, line width=0.5pt},
]
  \centering
  \includegraphics[width=\linewidth]{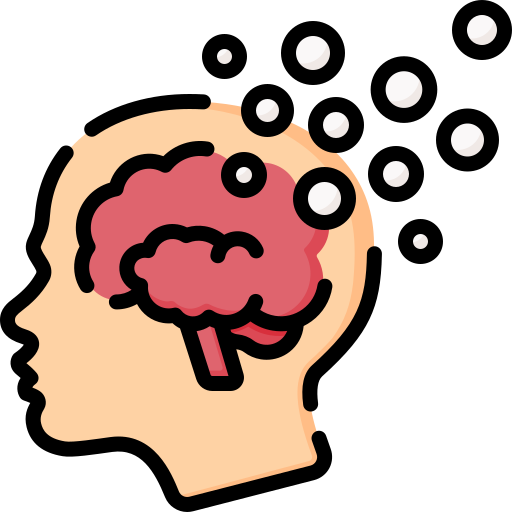}
  \tcblower
  \textit{Can VLA internal signals be transformed into self-evaluation metrics for action-generation reliability under a unified account of heterogeneous action-generation mechanisms, without relying on external supervision?}
\end{tcolorbox}

\textbf{The Present Work.}
We propose \ourmethod{} (Markov Attention Entropy), a white-box self-evaluation framework that converts a VLA's internal attention dynamics into reliability scores. Heterogeneous VLA architectures share a common action-generation structure: visual observations, language instructions, and State Input form the conditioning context, while latent action states evolve under this context toward the executable action.
The architectural difference lies in how this evolution is implemented.
In \textbf{Latent-Readout VLAs}, the policy evolves latent action representations and maps the final latent state to an executable action through a readout head, covering autoregressive discrete readout and continuous readout.
In \textbf{Latent-Refinement VLAs}, the policy maintains an action or action-trajectory representation and refines it through repeated update steps, covering flow-style refinement and diffusion-style generation.

\begin{figure*}[t]
    \centering
    \includegraphics[
        width=\textwidth,
        trim=0 5 0 50,
        clip
    ]{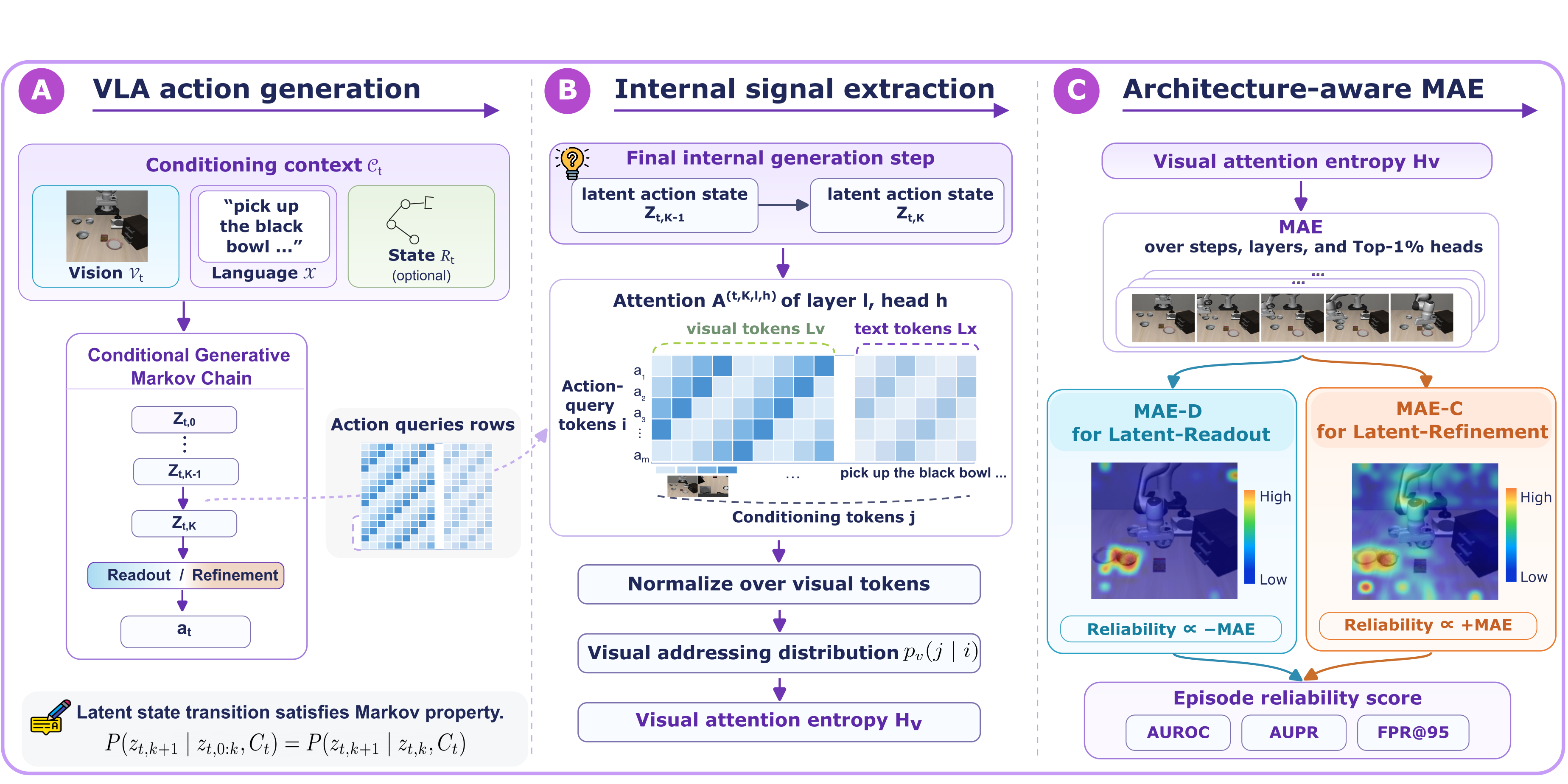}
    \caption{Unified Conditional Generative Markov Chain view of heterogeneous VLAs.
    Visual observations, language instructions, and State Input form the conditioning context, while latent action states evolve through architecture-specific transition kernels before the executable action is produced. \ourmethod{} reads attention entropy from this internal transition process and converts it into architecture-aware self-evaluation scores.}
    \label{fig:framework}
    \vspace{-1em}
\end{figure*}

This conditioned state evolution is captured by the Conditional Generative Markov Chain view illustrated in Figure~\ref{fig:framework}.
The latent action state is the Markov state of the internal generation process.
Conditioned on the current state and the conditioning context, the transition kernel determines how the next latent action state is formed, whether the update is implemented by autoregressive discrete readout, continuous readout, or flow-style refinement.
Attention records which visual and language tokens the current latent action state consults during this transition process.
Attention entropy measures this information routing and yields architecture-aware reliability scores.

 In Figure~\ref{fig:entropy-signal}, the paired attention maps show that successful and failed queries differ in how the latent action state addresses visual evidence, while the episode-level distributions show that visual-entropy scores separate reliable and unreliable executions more clearly than text-entropy scores on the same \subsetname{Reflect-Goal} episodes.
In the Markov formulation, visual attention entropy therefore measures how the transition kernel accesses visual perception when updating the latent action state.
For Latent-Readout VLAs, failed episodes show more diffuse visual addressing, and \maed scores the corresponding loss of concentration.
For Latent-Refinement VLAs, failed episodes show overly concentrated visual addressing at the final refinement step, and \maec scores the resulting loss of visual coverage before the action trajectory is returned.
Both metrics are oriented so that larger values indicate higher estimated reliability.

We evaluate \ourmethod on three open-source VLAs spanning the two action-generation families: \modelname{OpenVLA} and \modelname{OpenVLA-OFT} as Latent-Readout VLAs, and \modelname{QwenPI-Flow} as a Latent-Refinement VLA.
The evaluation uses \benchname{LIBERO-Reflect}, a 4,000-episode benchmark combining 2,000 standard episodes and 2,000 challenging episodes across four subsets.
The benchmark covers four capability axes, Goal Semantics, Object Binding, Spatial Grounding, and Composite Generalization, and evaluates whether a self-evaluation score ranks successful episodes above failed episodes.
Across \metricname{AUROC}, \metricname{AUPR}, and \metricname{FPR@95}, \ourmethod improves reliability ranking over all baselines without an external evaluator, repeated rollouts, or auxiliary model passes.
Our key contributions can be summarized as follows:
\begin{itemize}[leftmargin=2.2em,labelsep=0.55em,itemsep=0.35em,topsep=0.4em]
    \item[\textcolor{maePurple!85!black}{\ding{182}}] We formulate heterogeneous VLA action generation as a Conditional Generative Markov Chain, making the transition kernel the common object that links Latent-Readout VLAs and Latent-Refinement VLAs.
    \item[\textcolor{maePurple!85!black}{\ding{183}}] We introduce \ourmethod, a white-box self-evaluation framework that scores the visual-attention entropy of these transitions with architecture-aware orientations, \maed for Latent-Readout VLAs and \maec for Latent-Refinement VLAs.
    \item[\textcolor{maePurple!85!black}{\ding{184}}] We construct \benchname{LIBERO-Reflect}, a 4,000-episode benchmark combining 2,000 standard episodes and 2,000 challenging episodes across four subsets.
    \item[\textcolor{maePurple!85!black}{\ding{185}}] We show across three open-source VLAs that \ourmethod improves reliability ranking over black-box and white-box baselines without an external evaluator, repeated rollouts, or auxiliary model passes.
\end{itemize}

\section{Preliminary}

\subsection{The Conditional Generative Markov Chain}

We use a Conditional Generative Markov Chain to formalize the action-generation process of a VLA.
The chain describes the model's latent state evolution during action generation, not the physical robot-environment dynamics.
At timestep \(t\), the VLA evolves a sequence of latent action states \(z_{t,0:K}\in\mathcal{Z}\).
These states are hidden representations used for action generation and are not directly executed by the robot; the executable action \(a_t\in\mathcal{A}\), represented either as a single action or as an action chunk depending on the policy action head, is produced from the final latent action state, \(a_t=g_\theta(z_{t,K})\).
The chain is conditioned on the conditioning context \(C_t=(V_t,X,R_t)\in\mathcal{C}\), where \(V_t\) denotes visual representations, \(X\) denotes the language instruction, and \(R_t\) denotes optional State Input when used by the policy.
During the internal generation of \(a_t\), \(C_t\) is fixed, while \(z_{t,k}\) stores the complete action-generation state at internal generation step \(k\): the generated action-token prefix for autoregressive discrete readout, the latent action representation for continuous readout, or the current action trajectory for flow-style refinement.

\paragraph{Formal definition.}
We define the internal action-generation process as a conditional generative Markov chain $\mathcal{M}=(\mathcal{Z},\mathcal{C},K,\mathcal{P}_\theta)$, where $\mathcal{Z}$ is the latent action-state space, $\mathcal{C}$ is the conditioning space, $K$ is the number of internal generation steps, and $\mathcal{P}_\theta$ is the model's neural transition kernel.
At timestep \(t\), the internal process is \(z_{t,0}\rightarrow z_{t,1}\rightarrow \cdots \rightarrow z_{t,K}\rightarrow a_t\), where \(z_{t,k}\) is the latent action state at internal step \(k\), and \(a_t\) is the executable action.
The chain is specified by an initial distribution and an architecture-dependent transition kernel:
\begin{equation}
z_{t,0}\sim \rho_0(\cdot\mid C_t), \qquad z_{t,k+1}\sim \mathcal{P}_\theta(\cdot\mid z_{t,k},C_t), \quad k=0,\ldots,K-1.
\label{eq:markov-transitions}
\end{equation}
Here \(\rho_0\) is the initial latent action-state distribution, and \(\mathcal{P}_\theta\) is the neural transition kernel induced by the VLA architecture, such as autoregressive action-token updates, continuous readout, or flow-style refinement.
The probabilistic notation also covers deterministic updates as degenerate kernels.
Equivalently, the conditional joint distribution factorizes as
\begin{equation}
P_\theta(z_{t,0:K}\mid C_t)=\rho_0(z_{t,0}\mid C_t)\prod_{k=0}^{K-1}\mathcal{P}_\theta(z_{t,k+1}\mid z_{t,k},C_t),
\label{eq:markov-joint}
\end{equation}
The final action is \(a_t=g_\theta(z_{t,K})\), where \(g_\theta\) denotes the architecture-specific action head.

\paragraph{Markov property.}
The Markov property holds at the level of the \emph{complete} latent action state: given the current latent action state $z_{t,k}$ and the conditioning context $C_t$, the next latent action state $z_{t,k+1}$ does not depend additionally on earlier states $z_{t,0:k-1}$.
That is, \(P(z_{t,k+1}\mid z_{t,0:k},C_t)=P(z_{t,k+1}\mid z_{t,k},C_t)\).
Because \(z_{t,k}\) denotes the complete action-generation state, this formulation identifies the transition kernel as the common object shared by heterogeneous VLA action-generation.

\section{Methodology}

\subsection{Markov Attention Entropy}

Under the Conditional Generative Markov Chain interpretation, each transition \(z_{t,k-1}\rightarrow z_{t,k}\) is an information-routing step.
The transition kernel updates the latent action state by querying the conditioning context.
We use transformer attention to measure this routing behavior.
For internal generation step \(k\), layer \(\ell\), and head \(h\), let \(\mathbf{A}^{(t,k,\ell,h)}\in\mathbb{R}^{S_a\times S_{\mathrm{c}}}\) be the attention matrix from action-query tokens to conditioning context tokens, where \(S_a\) is the number of action-query tokens, and \(S_{\mathrm{c}}\) is the number of conditioning context tokens.
\ourmethod{} evaluates attention at \(k=K\), the final internal generation step before \(a_t\) is produced.
We decompose the key side into modality-specific token sets \(\mathcal{I}_v\) and \(\mathcal{I}_x\), corresponding to vision and language tokens.
For the visual modality, the normalized visual addressing distribution is
\begin{equation}
p_v^{(t,k,\ell,h)}(j\mid i)=\frac{A_{ij}^{(t,k,\ell,h)}}{\sum_{r\in\mathcal{I}_v}A_{ir}^{(t,k,\ell,h)}+\epsilon},\qquad j\in\mathcal{I}_v.
\end{equation}
The visual attention entropy is
\begin{equation}
H_v^{(t,k,\ell,h)}(i)=-\sum_{j\in\mathcal{I}_v}p_v^{(t,k,\ell,h)}(j\mid i)\log p_v^{(t,k,\ell,h)}(j\mid i).
\end{equation}
This entropy measures the uncertainty of visual addressing. A high value indicates that the action state distributes its visual attention broadly across many patches, while a low value indicates concentrated visual addressing over fewer patches. In the Conditional Generative Markov Chain view, \(H_v\) measures how the transition kernel \(\mathcal{P}_\theta\) accesses visual perception when updating the action state.

\paragraph{Episode \ourmethod{} score}
\label{sec:mae-score}
We convert the token-level visual entropy into an episode-level score by first averaging over execution steps and action queries:
\begin{equation}
E_{\ell,h}
=
\frac{1}{N}\sum_{t=1}^{N}
\frac{1}{S_a}\sum_{i=1}^{S_a}
H_v^{(t,K,\ell,h)}(i),
\end{equation}

where \(N\) is the number of execution steps in the episode, and \(E_{\ell,h}\) denotes the averaged visual entropy of head \(h\) in layer \(\ell\).
Let \(m\) be the number of selected heads per layer.
We define the episode-level \ourmethod score using an architecture-aware orientation \(\omega\in\{-1,+1\}\):
\begin{equation}
\begin{aligned}
\maemath_{\omega}^{(m)}(E)
&=
\frac{1}{L}\sum_{\ell=1}^{L}
\frac{1}{m}
\sum_{h\in \topk_{m}(-\omega E_{\ell,:})}
\omega E_{\ell,h}.
\end{aligned}
\end{equation}

Here, \(\topk_{m}(-\omega E_{\ell,:})\) selects the \(m\) heads corresponding to the architecture-aware entropy tail. For \(\omega=-1\), it selects the highest-entropy heads; for \(\omega=+1\), it selects the lowest-entropy heads. The factor \(\omega\) then orients both scores such that larger \ourmethod indicates higher estimated reliability.
The two \ourmethod metrics used in the experiments are \(\maedmath^{(m)}=\maemath_{-1}^{(m)}(E)\) and \(\maecmath^{(m)}=\maemath_{+1}^{(m)}(E)\).
\maed is used for Latent-Readout VLAs such as \modelname{OpenVLA} and \modelname{OpenVLA-OFT}, where reliable executions concentrate visual addressing before readout.
\maec is used for Latent-Refinement VLAs such as \modelname{QwenPI-Flow}, where reliable executions retain broader visual addressing at the final refinement step.

\section{\benchname{LIBERO-Reflect} Benchmark}
\label{sec:libero-reflect}

\paragraph{Matched benchmark construction.}
We construct \benchname{LIBERO-Reflect} for episode-level VLA self-evaluation by pairing standard episodes from \benchname{LIBERO} with challenging episodes from \benchname{LIBERO-PRO}. For each suite, we use all 10 tasks and execute every task under 50 non-identical initializations with small scene-state differences, yielding 500 standard episodes per suite. We retain the same task and initialization counts on the challenging side, where task language remains fixed while the placements of target and surrounding objects are swapped. Each subset therefore contains 500 standard and 500 challenging episodes; across four subsets, the benchmark contains 4,000 episodes in total, split evenly between 2,000 standard and 2,000 challenging episodes.

\paragraph{Capability subsets.}
The four subsets isolate complementary capabilities while preserving the same 1,000-episode composition.

\begin{tcbraster}[
  raster columns=2,
  raster equal height=rows,
  raster column skip=2mm,
  raster row skip=2mm,
]
  \begin{tcolorbox}[enhanced,colframe=maePurple!55,colback=white,boxrule=0.5pt,arc=1.5mm,left=1.5mm,right=1.5mm,top=1.2mm,bottom=1.2mm,fontupper=\footnotesize]
    \goalicon~\textbf{Goal Semantics}\hfill{\scriptsize\subsetname{Reflect-Goal}}\par\smallskip
    Goal-conditioned task understanding and action-type selection.\par\smallskip
    {\scriptsize\textcolor{maePurple!80!black}{500 standard + 500 challenging}}
  \end{tcolorbox}
  \begin{tcolorbox}[enhanced,colframe=maePurple!55,colback=white,boxrule=0.5pt,arc=1.5mm,left=1.5mm,right=1.5mm,top=1.2mm,bottom=1.2mm,fontupper=\footnotesize]
    \cubeicon~\textbf{Object Binding}\hfill{\scriptsize\subsetname{Reflect-Object}}\par\smallskip
    Target-object grounding and object--action binding.\par\smallskip
    {\scriptsize\textcolor{maePurple!80!black}{500 standard + 500 challenging}}
  \end{tcolorbox}
  \begin{tcolorbox}[enhanced,colframe=maePurple!55,colback=white,boxrule=0.5pt,arc=1.5mm,left=1.5mm,right=1.5mm,top=1.2mm,bottom=1.2mm,fontupper=\footnotesize]
    \routeicon~\textbf{Spatial Grounding}\hfill{\scriptsize\subsetname{Reflect-Spatial}}\par\smallskip
    Spatial-relation reasoning and layout-sensitive action adaptation.\par\smallskip
    {\scriptsize\textcolor{maePurple!80!black}{500 standard + 500 challenging}}
  \end{tcolorbox}
  \begin{tcolorbox}[enhanced,colframe=maePurple!55,colback=white,boxrule=0.5pt,arc=1.5mm,left=1.5mm,right=1.5mm,top=1.2mm,bottom=1.2mm,fontupper=\footnotesize]
    \compicon~\textbf{Composite Generalization}\hfill{\scriptsize\subsetname{Reflect-10}}\par\smallskip
    Mixed goals, objects, and spatial layouts from \benchname{LIBERO-10}.\par\smallskip
    {\scriptsize\textcolor{maePurple!80!black}{500 standard + 500 challenging}}
  \end{tcolorbox}
\end{tcbraster}

\paragraph{Episode-level assessment.}
Each method assigns one scalar reliability score to each rollout. Evaluation uses the rollout's actual \emph{simulator success flag} as the ground-truth label, rather than its nominal standard or challenging source. A stronger self-evaluation method should rank successful executions above failed ones.

\paragraph{Evaluation metrics.}
We report three complementary ranking metrics.

\begin{tcbraster}[raster columns=3,raster equal height=rows,raster column skip=2mm]
  \begin{tcolorbox}[enhanced,colframe=maePurple!55,colback=white,boxrule=0.5pt,arc=1.5mm,left=1.2mm,right=1.2mm,top=1.2mm,bottom=1.2mm,fontupper=\footnotesize]
    \centering\textbf{\metricname{AUROC}}~$\uparrow$\par\smallskip
    Global ranking quality across successful and failed episodes.
  \end{tcolorbox}
  \begin{tcolorbox}[enhanced,colframe=maePurple!55,colback=white,boxrule=0.5pt,arc=1.5mm,left=1.2mm,right=1.2mm,top=1.2mm,bottom=1.2mm,fontupper=\footnotesize]
    \centering\textbf{\metricname{AUPR}}~$\uparrow$\par\smallskip
    Precision-focused ranking quality under class imbalance.
  \end{tcolorbox}
  \begin{tcolorbox}[enhanced,colframe=maePurple!55,colback=white,boxrule=0.5pt,arc=1.5mm,left=1.2mm,right=1.2mm,top=1.2mm,bottom=1.2mm,fontupper=\footnotesize]
    \centering\textbf{\metricname{FPR@95}}~$\downarrow$\par\smallskip
    False-positive rate when 95\% of successful episodes are recalled.
  \end{tcolorbox}
\end{tcbraster}

Supporting construction statistics and nominal-to-actual label mappings are reported in \appref{app:libero-reflect-construction}, with dataset-source considerations in \appref{app:dataset-source}. The matched standard and challenging pools share suite organization, task count, initialization count, simulator, and rollout protocol. Because all metrics use realized simulator success rather than nominal source membership, \benchname{LIBERO-Reflect} evaluates reliability ranking within a matched mixed-difficulty pool rather than source classification.

\section{Experiments}

\label{sec:experiments}

In this section, we conduct extensive experiments to answer the following research questions: (\rqtag{1}) Can \ourmethod provide reliable self-evaluation across heterogeneous VLA architectures and various scenarios? (\rqtag{2}) Does \ourmethod introduce significant computational overhead? (\rqtag{3}) How sensitive is \ourmethod to its key components and hyperparameters?

\vspace{-0.5em}
\subsection{Experimental Setup}
\vspace{-0.5em}
\paragraph{VLA Backbones.}
We evaluate three representative open-source VLA policies on \benchname{LIBERO-Reflect}: \openvla, \openvlaoft, and \qwenpiflow. \openvla denotes the standard OpenVLA policy in Latent-Readout VLAs, with autoregressive discrete readout. \openvlaoft keeps the same OpenVLA backbone family and uses the OFT adaptation recipe with continuous readout. \qwenpiflow denotes a Qwen3-VL-based policy in Latent-Refinement VLAs, with a flow-style action module. Together, these policies pair two VLM backbone families with different action-generation mechanisms and training or adaptation regimes. This design tests whether \ourmethod transfers across action-generation architectures, VLM backbone families, and training sources. Detailed checkpoint names, vision-language backbones, action heads, and training sources are provided in \appref{app:model-configurations}.

\paragraph{Baselines.}
We compare against seven baselines covering black-box and white-box self-evaluation. The black-box group includes Random \citep{Bi2025CoTKineticsAT}, Verbal Confidence \citep{Kadavath2022LanguageM}, and Self-Consistency \citep{Wang2022SelfConsistencyIC}. The white-box group includes Maximum Softmax Probability \citep{si2023prompting}, Perplexity \citep{si2023prompting}, Entropy \citep{huang2023look}, and Length-normalized Entropy \citep{malinin2021uncertainty}. For all methods, we compute \metricname{AUROC}, \metricname{AUPR}, and \metricname{FPR@95} under the same episode labels. Random serves as a uniform-ranking lower bound. Because the evaluated VLAs use different action-generation interfaces, some baselines are technically not applicable to specific action heads; baseline definitions, applicability, OpenVLA-specific protocols, and the Verbal Confidence prompt are detailed in \appref{app:baseline-applicability}. The shared evaluation protocol is summarized in \appref{app:evaluation-settings}.

\paragraph{Parameter Settings.}
We obtain attention from the same policy forward pass used to generate actions, then compute the visual-entropy matrix \(E\) defined in the Markov Attention Entropy subsection. The main results use all layers and \topkname{1} head selection, selecting one architecture-aware critical head per layer. We report \maed for Latent-Readout VLAs and \maec for Latent-Refinement VLAs according to the entropy orientations above. \appref{app:ablation-source-data}, \appref{app:evaluation-settings}, and \appref{app:chat-templates} summarize exact ablation values, the common scoring protocol and model-specific layer/head settings, and policy input templates.

\begin{table}[t!]
  \centering
  \small
  \setlength{\tabcolsep}{1pt}
  \renewcommand{\arraystretch}{1.18}
  \resizebox{\textwidth}{!}{%
  \begin{tabular}{@{}>{\centering\arraybackslash}p{22mm}>{\raggedright\arraybackslash}p{20mm}*{12}{c}@{}}
    \toprule
    \multirow{2}{*}{Model} & \multirow{2}{*}{Method}
    & \multicolumn{3}{c}{\suiteheader{\goalicon}{Goal Semantics}{Reflect-Goal}}
    & \multicolumn{3}{c}{\suiteheader{\cubeicon}{Object Binding}{Reflect-Object}}
    & \multicolumn{3}{c}{\suiteheader{\routeicon}{Spatial Grounding}{Reflect-Spatial}}
    & \multicolumn{3}{c}{\suiteheader{\compicon}{Composite}{Reflect-10}} \\
    \cmidrule(lr){3-5} \cmidrule(lr){6-8} \cmidrule(lr){9-11} \cmidrule(lr){12-14}
    & & \mainmetricheader{AUROC}{\uparrow} & \mainmetricheader{AUPR}{\uparrow} & \mainmetricheader{FPR@95}{\downarrow}
      & \mainmetricheader{AUROC}{\uparrow} & \mainmetricheader{AUPR}{\uparrow} & \mainmetricheader{FPR@95}{\downarrow}
      & \mainmetricheader{AUROC}{\uparrow} & \mainmetricheader{AUPR}{\uparrow} & \mainmetricheader{FPR@95}{\downarrow}
      & \mainmetricheader{AUROC}{\uparrow} & \mainmetricheader{AUPR}{\uparrow} & \mainmetricheader{FPR@95}{\downarrow} \\
    \midrule
    \multicolumn{14}{c}{\textbf{\ding{182} Latent-Readout VLAs}} \\
    \midrule

    \multirow[c]{11}{22mm}[-0.55em]{\mainmodelcell{\llamaicon}{\modelname{OpenVLA}}}
    \baselinegroup{Black-box baselines}
    & Random & 48.55 & 39.27 & 95.36 & 47.85 & 35.98 & 95.87 & 48.97 & 39.49 & 94.82 & 52.51 & 28.88 & 93.92 \\
    & Verbal. Conf & 50.36 & 40.82 & 92.91 & 49.74 & 37.21 & 93.64 & 50.88 & 41.03 & 92.47 & 53.69 & 30.04 & 91.86 \\
    & Self-Cons.\draftrowmark & 56.82 & 41.28 & 84.90 & 58.41 & 42.76 & 82.35 & 57.94 & 44.15 & 87.80 & 53.64 & 27.20 & 90.64 \\
    \cmidrule(lr){2-14}
    \baselinegroup{White-box baselines}
    & MaxProb\draftrowmark & 54.72 & 40.58 & 88.62 & 55.83 & 40.28 & 86.74 & 55.61 & 42.37 & 90.10 & 53.28 & 27.05 & 91.48 \\
    & Perplexity\draftrowmark & 55.04 & 40.73 & 87.95 & 56.12 & 40.71 & 86.20 & 56.02 & 42.59 & 89.44 & 53.36 & 27.12 & 91.10 \\
    & Entropy\draftrowmark & 56.10 & 41.05 & 86.80 & 57.48 & 41.64 & 84.75 & 57.17 & 43.26 & 88.37 & 53.82 & 27.30 & 90.38 \\
    & Length Ent.\draftrowmark & 56.38 & 41.19 & 86.32 & 58.02 & 42.05 & 84.12 & 57.52 & 43.58 & 87.90 & 54.02 & 27.38 & 89.96 \\
    \maeboxrule

    & \mainmaemethod{\maed{}}{16}
    & \metricgainunder{59.56}{+22.7\%} & \metricgainunder{40.67}{+3.6\%} & \metricgainunder{71.52}{$\downarrow$25.0\%}
    & \metricgainunder{79.64}{+66.4\%} & \metricgainunder{58.98}{+63.9\%} & \metricgainunder{48.81}{$\downarrow$49.1\%}
    & \metricgainunder{63.99}{+30.7\%} & \metricgainunder{49.67}{+25.8\%} & \metricgainunder{84.81}{$\downarrow$10.6\%}
    & \metricgainunder{53.21}{+1.3\%} & \metricgainplain{26.56}{-8.0\%} & \metricgainunder{87.43}{$\downarrow$6.9\%} \\

    & \mainmaemethod{\maed{}}{1}
    & \metricgain{63.94}{+31.7\%} & \metricgain{43.23}{+10.1\%} & \metricgain{61.92}{$\downarrow$35.1\%}
    & \metricgain{90.97}{+90.1\%} & \metricgain{75.88}{+110.9\%} & \metricgain{28.30}{$\downarrow$70.5\%}
    & \metricgain{66.86}{+36.5\%} & \metricgain{50.74}{+28.5\%} & \metricgain{75.79}{$\downarrow$20.1\%}
    & \metricgain{54.74}{+4.2\%} & \metricgain{30.45}{+5.4\%} & \metricgain{86.46}{$\downarrow$7.9\%} \\
    \maeboxrule

    \cmidrule(lr){1-14}

    \multirow[c]{5}{22mm}[-0.55em]{\mainmodelcell{\llamaicon}{\modelname{OpenVLA-OFT}}}
    \baselinegroup{Black-box baselines}
    & Random & 47.41 & 50.24 & 95.51 & 48.87 & 49.86 & 95.56 & 47.97 & 48.74 & 94.96 & 51.01 & 48.44 & 96.58 \\
    & Verbal. Conf & 49.12 & 51.73 & 92.84 & 50.68 & 51.08 & 93.11 & 49.63 & 50.26 & 92.75 & 52.45 & 50.01 & 93.89 \\
    \maeboxrule

    & \mainmaemethod{\maed{}}{16}
    & \metricgainunder{91.84}{+93.7\%} & \metricgainunder{94.50}{+88.1\%} & \metricgainunder{7.94}{$\downarrow$91.7\%}
    & \metricgainunder{75.96}{+55.4\%} & \metricgainunder{82.86}{+66.2\%} & \metricgainunder{64.03}{$\downarrow$33.0\%}
    & \metricgainunder{91.71}{+91.2\%} & \metricgainunder{89.85}{+84.3\%} & \metricgainunder{42.74}{$\downarrow$55.0\%}
    & \metricgainunder{77.13}{+51.2\%} & \metricgainunder{62.10}{+28.2\%} & \metricgainunder{45.06}{$\downarrow$53.3\%} \\

    & \mainmaemethod{\maed{}}{1}
    & \metricgain{97.34}{+105.3\%} & \metricgain{96.14}{+91.4\%} & \metricgain{7.14}{$\downarrow$92.5\%}
    & \metricgain{80.56}{+64.8\%} & \metricgain{81.10}{+62.7\%} & \metricgain{63.71}{$\downarrow$33.3\%}
    & \metricgain{92.63}{+93.1\%} & \metricgain{93.15}{+91.1\%} & \metricgain{39.31}{$\downarrow$58.6\%}
    & \metricgain{78.57}{+54.0\%} & \metricgain{64.08}{+32.3\%} & \metricgain{45.06}{$\downarrow$53.3\%} \\
    \maeboxrule

    \midrule
    \multicolumn{14}{c}{\textbf{\ding{183} Latent-Refinement VLAs}} \\
    \midrule

    \multirow[c]{5}{22mm}[-0.55em]{\mainmodelcell{\qwenicon}{\modelname{QwenPI-Flow}}}
    \baselinegroup{Black-box baselines}
    & Random & 47.58 & 47.93 & 95.91 & 48.75 & 48.40 & 95.10 & 48.11 & 50.15 & 94.82 & 50.81 & 48.95 & 96.53 \\
    & Verbal. Conf & 49.44 & 49.51 & 93.42 & 50.31 & 49.82 & 92.68 & 49.96 & 51.62 & 92.31 & 52.18 & 50.47 & 93.77 \\
    \maeboxrule

    & \mainmaemethod{\maec{}}{20}
    & \metricgainunder{62.63}{+31.6\%} & \metricgainunder{59.16}{+23.4\%} & \metricgainunder{80.93}{$\downarrow$15.6\%}
    & \metricgainunder{65.92}{+35.2\%} & \metricgainunder{63.43}{+31.1\%} & \metricgainunder{81.59}{$\downarrow$14.2\%}
    & \metricgainunder{66.89}{+39.0\%} & \metricgainunder{64.70}{+29.0\%} & \metricgainunder{73.50}{$\downarrow$22.5\%}
    & \metricgainunder{68.91}{+35.6\%} & \metricgainunder{61.92}{+26.5\%} & \metricgainunder{74.52}{$\downarrow$22.8\%} \\

    & \mainmaemethod{\maec{}}{1}
    & \metricgain{80.57}{+69.3\%} & \metricgain{80.01}{+66.9\%} & \metricgain{60.12}{$\downarrow$37.3\%}
    & \metricgain{75.94}{+55.8\%} & \metricgain{76.48}{+58.0\%} & \metricgain{81.18}{$\downarrow$14.6\%}
    & \metricgain{84.80}{+76.3\%} & \metricgain{85.46}{+70.4\%} & \metricgain{68.53}{$\downarrow$27.7\%}
    & \metricgain{79.52}{+56.5\%} & \metricgain{76.24}{+55.8\%} & \metricgain{55.60}{$\downarrow$42.4\%} \\
    \maeboxrule

    \bottomrule
  \end{tabular}
  }
  \caption{Main self-evaluation results on \benchname{LIBERO-Reflect}. Applicable baselines are grouped as black-box or white-box methods. Purple rules highlight \ourmethod rows, which report the score and relative change against \texttt{Random}; for \metricname{FPR@95}, the relative value is the reduction rate. Bold \topkname{1} values and underlined half-head values improve over \texttt{Random} within the same model, subset, and metric. \(^{\dagger}\) Token-statistic and Self-Consistency baselines are reported only for \modelname{OpenVLA}, whose autoregressive discrete action interface exposes action-token probabilities and sampled action-token sequences. Additional applicability details are provided in \appref{app:baseline-applicability}.}
  \label{tab:vla-main-results}
  \vspace{-2em}
\end{table}

\vspace{-0.5em}
\subsection{Main Results (\rqtag{1})}
\vspace{-0.5em}

\takeaway{202}{\ourmethod improves over black-box and white-box self-evaluation baselines.}
Table~\ref{tab:vla-main-results} provides a comprehensive analysis of \ourmethod under the full black-box and white-box baseline protocol across three VLA backbones and four \benchname{LIBERO-Reflect} subsets. \ourmethod directly scores the internal visual-attention entropy of latent action generation. Against the Random lower bound, \maed improves \openvla by up to 90.1\% \metricname{AUROC} on Object Binding and reduces \metricname{FPR@95} by 70.5\%; \maed improves \openvlaoft by 105.3\% \metricname{AUROC} on Goal Semantics; and \maec improves \qwenpiflow by 76.3\% \metricname{AUROC} on Spatial Grounding.
For the discrete \openvla policy, the output-statistic and Self-Consistency rows instantiate baselines that depend on action-token probabilities or sampled action-token agreement, while the continuous-action and flow-action policies cannot expose the same random variables without changing the estimator. The OpenVLA-specific baseline protocols are detailed in \appref{app:baseline-applicability}.

\takeaway{203}{\ourmethod transfers across model architectures, initialization families, and task families.}
The three evaluated policies differ in base VLMs, perception stacks, training data, and action-generation mechanisms. \openvla and \openvlaoft instantiate Latent-Readout VLAs with different readout recipes. \qwenpiflow instantiates Latent-Refinement VLAs with a flow-style action module. The gains on Goal Semantics, Object Binding, and Spatial Grounding show that the signal follows the action-condition coupling structure across distinct capability axes.

\takeaway{204}{\ourmethod yields a shared execution-correctness signal across heterogeneous tasks.}
\subsetname{Reflect-10} pools diverse goals, objects, and layouts from \benchname{LIBERO-10} into a single evaluation split. On this heterogeneous task mixture, \openvlaoft reaches 78.57 \metricname{AUROC} and \qwenpiflow reaches 79.52 \metricname{AUROC}. The entropy signal thus functions as a cross-task execution monitor, discriminating successful from failed rollouts as task semantics and scene configurations vary jointly.

\begin{figure}[t!]
  \centering
  \begin{subfigure}[t]{0.48\textwidth}
    \centering
    \includegraphics[width=\linewidth]{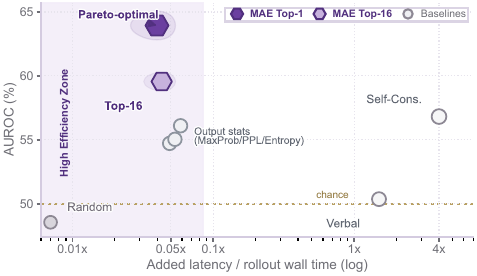}
    \phantomcaption
    \label{fig:efficiency-pareto}
  \end{subfigure}
  \hfill
  \begin{subfigure}[t]{0.48\textwidth}
    \centering
    \includegraphics[width=\linewidth]{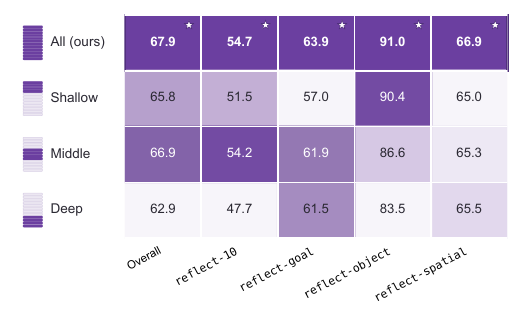}
    \phantomcaption
    \label{fig:openvla-layer-ablation}
  \end{subfigure}
  \caption{Efficiency--reliability Pareto analysis and layer-band ablation for \ourmethod. The left panel reports added evaluation latency normalized by one rollout wall time. The right panel reports absolute \metricname{AUROC} for \maed{} with \topkname{1}; colors are normalized within each column by closeness to the all-layer result, stars mark the best layer band, and the left schematic indicates the layer region.}
  \vspace{-1em}
  \label{fig:efficiency-and-layer-ablation}
\end{figure}

\subsection{Cost Analysis (\rqtag{2})}

\takeaway{205}{\ourmethod achieves the highest performance in the high-efficiency zone, establishing itself as the Pareto-optimal solution.} Figure~\ref{fig:efficiency-pareto} visualizes the efficiency-reliability Pareto view. While baselines like Self-Consistency or Verbal Confidence require multiple sampled action-token generations or external models (incurring $\ge 1\times$ latency overhead), \ourmethod reuses internal attention maps and adds minimal computation ($<0.1\times$ overhead; detailed hardware profiling on NVIDIA H100 is provided in \appref{app:cost-analysis-details}).

\vspace{-0.5em}
\subsection{Framework Analysis  (\rqtag{3})}
\vspace{-0.5em}
\begin{wrapfigure}[16]{r}{0.5\textwidth}
  \vspace{-1.0\baselineskip}
  \centering
  \includegraphics[width=\linewidth]{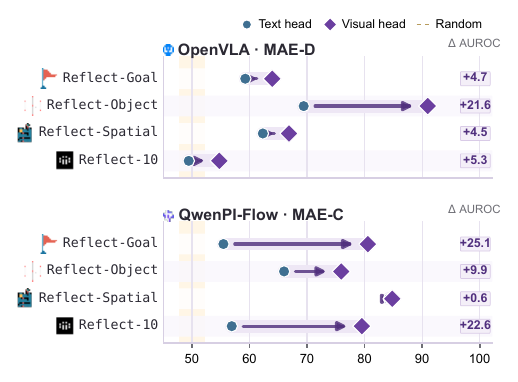}
  \caption{Text-head versus visual-head self-evaluation signals. \openvla uses \maed{}, and \qwenpiflow uses \maec{}. Each connector reports \(\Delta\)\metricname{AUROC} = V - T. Positive gaps across all subsets indicate that text-side entropy is not a stable substitute for visual information.}
  \label{fig:text-visual-head-comparison}
  \vspace{-0.75\baselineskip}
\end{wrapfigure}
We ablate the key design choices in \ourmethod from the following perspectives:

\paragraph{Text-head versus visual-head signals.}
We first test whether the reliability signal can be recovered from the language side alone. Figure~\ref{fig:text-visual-head-comparison} compares the text-head \ourmethod{}~\topkname{1} score against the visual-attention \ourmethod{}~\topkname{1} score used in Table~\ref{tab:vla-main-results}; exact values are reported in \appref{app:ablation-source-data}. The visual head consistently improves \metricname{AUROC}, with especially large gains on \openvla Object Binding and \qwenpiflow Goal Semantics and \subsetname{Reflect-10}. This shows that language-head entropy can reflect text-side decisiveness, but it does not reliably capture whether the action latent has gathered the visual evidence needed for manipulation. The visual head is therefore the appropriate signal for self-evaluating action generation.

\begin{figure}[t!]
  \centering
  \includegraphics[width=\textwidth]{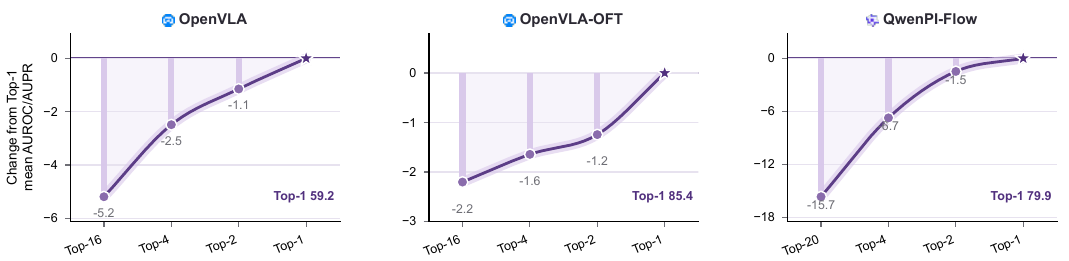}
  \caption{\topkname{m} head selection ablation. Panels order selected-head settings from many heads to \topkname{1} and use independent y-axis scales. Values are percentage-point changes from the \topkname{1} ranking score, computed as the mean of \metricname{AUROC} and \metricname{AUPR} over the four \benchname{LIBERO-Reflect} subsets. Exact values are reported in \appref{app:ablation-source-data}.}
  \label{fig:topk-head-selection}
  \vspace{-2em}
\end{figure}

\paragraph{Impact of head selection.}
We evaluate how the number of selected heads affects \ourmethod by varying \topkname{m}. Figure~\ref{fig:topk-head-selection} summarizes the full results by plotting each setting's drop from \topkname{1} in mean ranking quality, averaged over \metricname{AUROC} and \metricname{AUPR} across the four subsets; exact values are reported in \appref{app:ablation-source-data}. \topkname{1} gives the most consistent ranking performance across all three models. It also uses the smallest selected-head set, reducing entropy aggregation cost and limiting the contribution of low-signal heads. This supports \topkname{1} as the default head-selection setting.

\paragraph{Impact of layer selection.}
To assess the effect of model depth, we compare the default all-layer aggregation with three layer-restricted variants: shallow, middle, and deep. The experiment keeps the same \maed score with \topkname{1} and changes only the layer band used to aggregate \(E_{\ell,h}\). Figure~\ref{fig:openvla-layer-ablation} reports absolute \metricname{AUROC} values while coloring each column by closeness to the all-layer result, showing that all-layer aggregation gives the best overall \metricname{AUROC} and remains competitive across task families. This result is consistent with the layered function of transformer VLAs: earlier layers encode visual grounding and token-level perception, middle layers support cross-modal binding, and deeper layers are closer to action readout. Combining all layers provides a fuller internal signal and avoids introducing task- or architecture-specific layer-band hyperparameters. 

\paragraph{Run-to-run stability.}

\begin{wraptable}[14]{r}{0.50\textwidth}
  \vspace{0\baselineskip}
  \centering
  \scriptsize
  \setlength{\tabcolsep}{2.2pt}
  \renewcommand{\arraystretch}{1.12}
  \resizebox{\linewidth}{!}{%
  \begin{tabular}{llccc}
    \toprule
    \textbf{Benchmark subset} & \textbf{Method} & \metricname{AUROC}$\uparrow$ & \metricname{AUPR}$\uparrow$ & \metricname{FPR@95}$\downarrow$ \\
    \midrule
    \multirow{2}{*}{\makecell[l]{\subsetname{Reflect-Goal}\\[-0.18em]\scriptsize Goal Semantics}} & Random & $47.75{\pm}0.81$ & $48.12{\pm}0.79$ & $95.98{\pm}0.63$ \\
    & \maemethodcell{\maec{}~\topkname{1}} & $\mathbf{80.72{\pm}0.79}$ & $\mathbf{80.14{\pm}0.78}$ & $\mathbf{60.15{\pm}1.28}$ \\
    \midrule
    \multirow{2}{*}{\makecell[l]{\subsetname{Reflect-Object}\\[-0.18em]\scriptsize Object Binding}} & Random & $48.91{\pm}0.79$ & $48.56{\pm}0.79$ & $95.18{\pm}0.64$ \\
    & \maemethodcell{\maec{}~\topkname{1}} & $\mathbf{76.12{\pm}0.85}$ & $\mathbf{76.64{\pm}0.83}$ & $\mathbf{81.13{\pm}1.27}$ \\
    \midrule
    \multirow{2}{*}{\makecell[l]{\subsetname{Reflect-Spatial}\\[-0.18em]\scriptsize Spatial Grounding}} & Random & $48.30{\pm}0.86$ & $50.31{\pm}0.86$ & $94.90{\pm}0.62$ \\
    & \maemethodcell{\maec{}~\topkname{1}} & $\mathbf{84.95{\pm}0.77}$ & $\mathbf{85.60{\pm}0.77}$ & $\mathbf{68.46{\pm}1.32}$ \\
    \midrule
    \multirow{2}{*}{\makecell[l]{\subsetname{Reflect-10}\\[-0.18em]\scriptsize Composite Generalization}} & Random & $50.99{\pm}0.82$ & $49.13{\pm}0.83$ & $96.63{\pm}0.65$ \\
    & \maemethodcell{\maec{}~\topkname{1}} & $\mathbf{79.71{\pm}0.80}$ & $\mathbf{76.41{\pm}0.80}$ & $\mathbf{55.59{\pm}1.26}$ \\
    \bottomrule
  \end{tabular}%
  }
  \caption{Run-to-run stability of \modelname{QwenPI-Flow} with \maec{}~\topkname{1} across all four subsets. Values are percentages reported as sample mean $\pm$ sample standard deviation over five independent simulator seeds.}
  \label{tab:repeated-run-stability}
  \vspace{-0.7\baselineskip}
\end{wraptable}
We repeat the full \modelname{QwenPI-Flow} rollout evaluation using five independent simulator seeds. Each seed changes the scene initialization and stochastic rollout trajectory while keeping the policy checkpoint, task set, and evaluation protocol fixed. Table~\ref{tab:repeated-run-stability} shows consistent gains from \maec{}~\topkname{1} across all four subsets, with standard deviations no larger than 1.32 percentage points, including for the tail-sensitive \metricname{FPR@95}. The improvements are therefore stable across random trials rather than being driven by a favorable single run. Under a two-sided Student-$t$ interval with four degrees of freedom, the largest 95\% confidence-interval half-width is 1.64 percentage points. Because \metricname{AUPR} depends on positive prevalence, comparisons retain the same episode composition within each subset.

\begin{figure}[t!]
  \centering
  \includegraphics[width=0.6\textwidth]{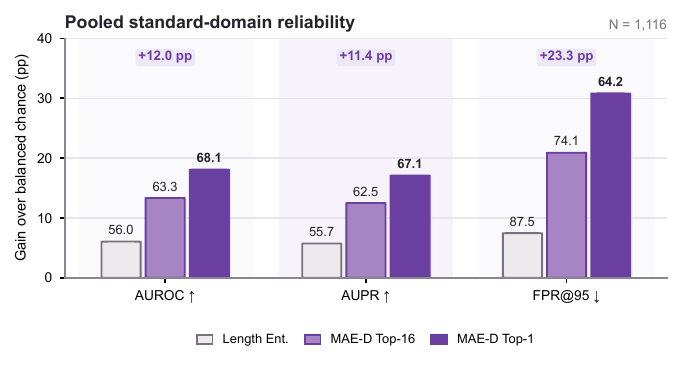}
  \caption{Pooled standard-only balanced results for \openvla. Bar heights report improvement over the balanced random expectation: score minus 50 for \metricname{AUROC}/\metricname{AUPR}, and 95 minus score for \metricname{FPR@95}. Labels above bars give the original metric values. Purple callouts report the gain of \maed{}~\topkname{1} over Length-normalized Entropy.}
  \label{fig:standard-only-balanced}
  \vspace{-1em}
\end{figure}

\paragraph{\ourmethod retains within-domain execution reliability.}

To further assess execution-level reliability under a shared environment distribution, we evaluate \openvla using only standard LIBERO episodes. We retain all 558 natural policy failures and sample 558 subset-matched successes, yielding a balanced set of 1,116 episodes with 50\% success prevalence. Under this setting, \maed{}~\topkname{1} improves over Length-normalized Entropy by 12.02 points in pooled \metricname{AUROC}, 11.36 points in \metricname{AUPR}, and 23.34 points in \metricname{FPR@95} reduction. The same ordering holds across all four subsets; on Object Binding, \maed{}~\topkname{1} reaches 88.75 \metricname{AUROC}, 86.30 \metricname{AUPR}, and 30.15 \metricname{FPR@95}. These results show that MAE preserves reliable execution ranking even when successful and failed episodes are drawn from the same standard-domain distribution. Full construction details and subset-level results are provided in \appref{app:standard-only-balanced}.

\vspace{-1em}
\section{Conclusion}
\vspace{-1em}
In this work, we observe that internal visual modality entropy exhibits consistent distinctions between successful and failed tasks across heterogeneous VLAs. We presented \ourmethod{}, a white-box framework for VLA self-evaluation from internal attention dynamics. We introduce \benchname{LIBERO-Reflect}, a 4,000-episode benchmark combining 2,000 standard episodes and 2,000 challenging episodes across four subsets. Experiments on \benchname{LIBERO-Reflect} show that \ourmethod{} improves reliability ranking without external evaluators, establishing itself as the Pareto-optimal solution.

\FloatBarrier
\bibliographystyle{plainnat}
\bibliography{custom}

\clearpage
\beginappendix
\renewcommand{\maemethodcell}[1]{#1}
\section{Related Work}
\label{app:related-work}

\paragraph{Vision-Language-Action Models.}
VLAs translate visual observations and language instructions into output actions, building on progress in large-scale robot learning and multimodal modeling. Early generalist policies such as \modelname{RT-1} demonstrate scalable transformer-based control from real-world robot trajectories, while \modelname{VIMA} studies robot manipulation conditioned on multimodal task specifications \citep{brohan2023rt1roboticstransformerrealworld,jiang2023vimageneralrobotmanipulation}. \modelname{PaLM-E} incorporates visual observations into a pretrained Large Language Model (LLM), and \modelname{RT-2} connects web-scale vision-language pretraining with action generation through action-token prediction \citep{pmlr-v202-driess23a,pmlr-v229-zitkovich23a}. Large-scale datasets and open policies further support generalization across tasks and robot embodiments, including \modelname{Open X-Embodiment}, \modelname{RoboCat}, and \modelname{Octo} \citep{embodimentcollaboration2025openxembodimentroboticlearning,bousmalis2023robocatselfimprovinggeneralistagent,Ghosh-RSS-24}. Beyond these policy-scale developments, continuous generative policies such as \modelname{Diffusion Policy} and efficient action representations such as \modelname{FAST} expand the available designs for action generation \citep{chi2024diffusionpolicyvisuomotorpolicy,pertsch2025fastefficientactiontokenization}. Together, these studies establish a broad family of policies that share visual observations and language instructions as conditions, but differ in how actions are generated.

\paragraph{Self-Evaluation in Large Language Models.}
Self-evaluation in LLMs studies whether a model can estimate the reliability of its own generated outputs. Calibration-based work evaluates whether model confidence is aligned with answer correctness, including probability-based calibration and uncertainty expressed through natural-language confidence statements \citep{jiang-etal-2021-know,lin2022teachingmodelsexpressuncertainty,tian2023justaskcalibrationstrategies}. A second line of work uses repeated generation to measure agreement among candidate responses: self-consistency improves reasoning by aggregating multiple sampled outputs, while \modelname{SelfCheckGPT} detects hallucinations from inconsistencies among sampled passages \citep{Wang2022SelfConsistencyIC,manakul-etal-2023-selfcheckgpt}. Moving beyond surface-level agreement, semantic uncertainty groups generations according to their meanings, and semantic entropy uses this semantic-level uncertainty to identify unreliable generations and confabulations \citep{kuhn2023semanticuncertaintylinguisticinvariances,article}. Recent probing studies additionally indicate that internal hidden representations can expose reliability-related signals without requiring full repeated sampling \citep{azaria2023internalstatellmknows,kossen2024semanticentropyprobesrobust}. These works provide important foundations for self-evaluation, but their outputs are language generations rather than VLAs' output actions conditioned on visual observations.

\paragraph{Self-Evaluation in Vision-Language-Action Models.}
Self-evaluation for VLAs must assess the reliability of action generation under visual observations and language instructions. Recent studies evaluate VLA uncertainty or confidence from generated output actions, including uncertainty-quality evaluation and calibration methods for VLA policies \citep{valle2025evaluatinguncertaintyqualityvisual,zollo2025confidencecalibrationvisionlanguageactionmodels}. Other work detects or reasons over failures through additional learned components: \modelname{AHA} uses a VLM to identify and explain robotic manipulation failures, and \modelname{RoboMonkey} introduces test-time sampling and verification for VLA action generation \citep{duan2024ahavisionlanguagemodeldetectingreasoning, kwok2025robomonkeyscalingtesttimesampling}. Complementary studies begin to examine reliability-related signals already present inside VLAs, showing that internal signals can reflect controllable behavior, path deviation, or model limitations \citep{haeon2025mechanisticinterpretabilitysteeringvisionlanguageaction,jeong2026visionlanguageactionmodelattentionheads,wang2026vlaknowslimits}. Different from methods based on external evaluators, supervised failure detectors, or repeated action sampling, \ourmethod{} directly converts internal visual attention entropy into episode-level reliability scores for heterogeneous VLA action-generation mechanisms.

\section{Model Configuration Details}
\label{app:model-configurations}
\label{sec:appendix}

Table~\ref{tab:model-configurations} maps the model names used in the experiments to the concrete evaluated checkpoints and training configurations. The selected policies deliberately pair two VLM backbone families with different action-generation mechanisms and training or adaptation regimes. This design supports the claim that \ourmethod is effective across action-generation architectures, VLM backbone families, and training sources.

\begin{table*}[t!]
  \centering
  \small
  \setlength{\tabcolsep}{2pt}
  \renewcommand{\arraystretch}{1.18}
  \begin{tabular}{p{0.18\textwidth}p{0.15\textwidth}p{0.20\textwidth}p{0.23\textwidth}p{0.20\textwidth}}
    \toprule
    \textbf{Policy} & \textbf{Action-Generation Family} & \textbf{VLM and Perception Stack} & \textbf{Training / Adaptation Signal} & \textbf{Diversity Axis} \\
    \midrule
    \modelcell{\llamaicon}{\modelname{OpenVLA}} & Latent-Readout VLAs & OpenVLA-7B VLM with prism-dinosiglip-224px; fused DINOv2 + SigLIP visual encoder; Llama 2 7B language backbone & official openvla/openvla-7b; tokenized action prediction trained on Open X-Embodiment & Llama/OpenVLA family; large-scale action-token training; tests autoregressive discrete readout behavior \\
    \midrule
    \modelcell{\llamaicon}{\modelname{OpenVLA-OFT}} & Latent-Readout VLAs & OpenVLA-7B VLM with the same DINOv2 + SigLIP visual base and Llama 2 7B language backbone & official OpenVLA-OFT checkpoints/code; efficient OFT adaptation with continuous actions, action chunking, and L1 regression & Same backbone family as OpenVLA with a continuous readout recipe; tests adaptation robustness \\
    \midrule
    \modelcell{\qwenicon}{\modelname{QwenPI-Flow}} & Latent-Refinement VLAs & Qwen3-VL-4B-Instruct with Qwen3-VL vision stack plus dinov2\_vits14 in the StarVLA config & StarVLA/Qwen3-VL-PI-LIBERO-4in1; QwenPI policy with DiT-B, 7-DoF actions, horizon 8, and data\_mix=libero\_all & Qwen3-VL family; flow-style refinement; LIBERO-specific training mix; tests cross-family and cross-generation generality \\
    \bottomrule
\end{tabular}
  \caption{Detailed configurations for the three VLA policies evaluated in the main experiments. The table highlights architectural, family-level, and training-source diversity.}
  \label{tab:model-configurations}
\end{table*}

\section{\benchname{LIBERO-Reflect} Construction Details}
\label{app:libero-reflect-construction}

\begin{figure*}[p]
  \centering
  \includegraphics[width=\textwidth]{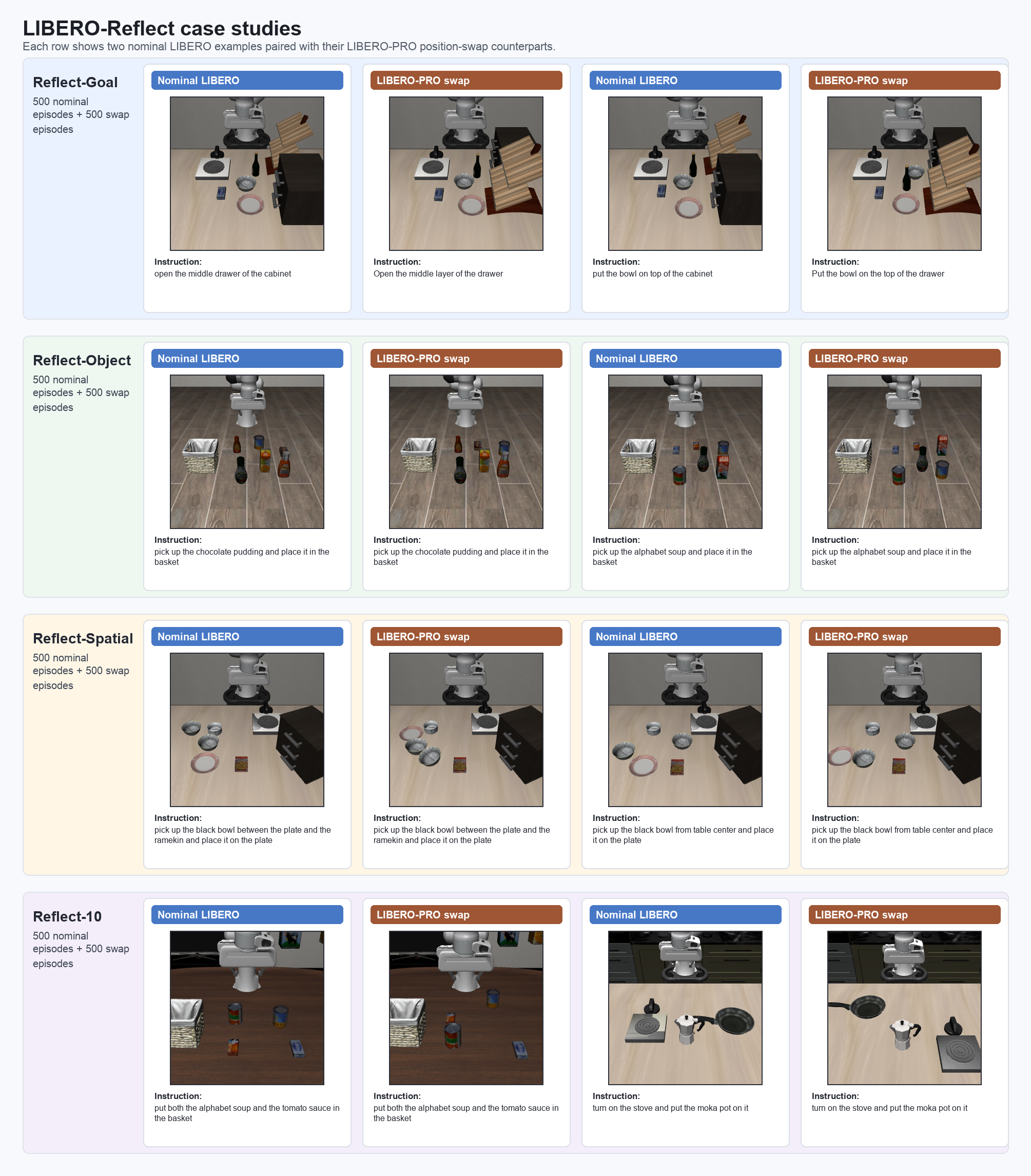}
  \caption{Case studies from the \benchname{LIBERO-Reflect} construction. Each row corresponds to one subset and shows representative standard episodes paired with challenging episodes. The instruction is printed below each panel, yielding 16 data points across the four subsets.}
  \label{fig:libero-reflect-case-study}
\end{figure*}

This appendix documents the construction evidence for \benchname{LIBERO-Reflect}. Table~\ref{tab:libero-reflect-composition} fixes the benchmark composition: each subset contains 500 standard episodes and 500 challenging episodes, yielding 1,000 episodes per subset and 4,000 episodes in total. The same table reports policy success rates on the standard side. Table~\ref{tab:libero-pro-negative-success} reports diagnostic success rates for the challenging episodes sampled from \benchname{LIBERO-PRO}. Figure~\ref{fig:libero-reflect-case-study} complements the tables with 16 visual case-study panels from all four subsets.
The challenging episodes preserve the original task language while swapping the placements of target objects and surrounding objects.

\paragraph{Nominal splits versus actual evaluation labels.}
The 2,000 standard and 2,000 challenging episodes define the \emph{nominal} composition of \benchname{LIBERO-Reflect}, designed to ensure sufficient exposure to both solvable standard scenes and difficult PRO scenes. However, depending on the evaluated policy's capability, nominal standard episodes may occasionally fail (as shown in Table~\ref{tab:libero-reflect-composition}), and nominal challenging episodes may occasionally succeed (as shown in Table~\ref{tab:libero-pro-negative-success}). Because self-evaluation must assess whether the policy \emph{actually} succeeded or failed on a given execution, we do not use the nominal dataset split as the ground-truth label for evaluation. Instead, the ground-truth label for computing all self-evaluation metrics (\metricname{AUROC}, \metricname{AUPR}, \metricname{FPR@95}) is strictly defined by the actual simulator success flag of that specific rollout. This ensures that our evaluation rigorously reflects the true correctness of the generated actions, avoiding any label noise introduced by policy capability variations.

\begin{table*}[t!]
  \centering
  \small
  \setlength{\tabcolsep}{3pt}
  \renewcommand{\arraystretch}{1}
  \resizebox{\textwidth}{!}{%
  \begin{tabular}{lccccc}
    \toprule
    \textbf{Component}
    & \suiteheader{\goalicon}{Goal Semantics}{Reflect-Goal}
    & \suiteheader{\routeicon}{Spatial Grounding}{Reflect-Spatial}
    & \suiteheader{\compicon}{Composite}{Reflect-10}
    & \suiteheader{\cubeicon}{Object Binding}{Reflect-Object}
    & \textbf{Overall} \\
    \midrule
    Task count & 10 & 10 & 10 & 10 & 40 \\
    Standard episodes & 500 & 500 & 500 & 500 & 2,000 \\
    Challenging episodes & 500 & 500 & 500 & 500 & 2,000 \\
    Total episodes & 1,000 & 1,000 & 1,000 & 1,000 & 4,000 \\
    \midrule
    \multicolumn{6}{c}{\textbf{\ding{182} Standard-side policy success rate (\%)}} \\
    \midrule
    \modelcell{\qwenicon}{\modelname{QwenPI-Flow}} & 96.8 & 93.8 & 94.4 & 98.0 & 95.8 \\
    \modelcell{\llamaicon}{\modelname{OpenVLA-OFT}} & 97.2 & 93.6 & 94.8 & 99.6 & 96.3 \\
    \modelcell{\llamaicon}{\modelname{OpenVLA}} & 79.2 & 79.8 & 55.2 & 74.2 & 72.1 \\
    \bottomrule
  \end{tabular}
  }%
  \caption{\benchname{LIBERO-Reflect} composition and standard-side policy success rates. The upper block reports the number of tasks and episodes used to form the benchmark. The lower block reports policy success rates on the standard side. Overall success is the mean over four equally sized suites. \modelname{QwenPI-Flow} corresponds to the Qwen3-VL-PI-LIBERO-4in1 checkpoint described in Table~\ref{tab:model-configurations}.}
  \label{tab:libero-reflect-composition}
\end{table*}

\begin{table*}[t!]
  \centering
  \small
  \setlength{\tabcolsep}{3pt}
  \renewcommand{\arraystretch}{1.18}
  \resizebox{\textwidth}{!}{%
  \begin{tabular}{lccccc}
    \toprule
    \textbf{Policy}
    & \suiteheader{\goalicon}{Goal Semantics}{Reflect-Goal}
    & \suiteheader{\routeicon}{Spatial Grounding}{Reflect-Spatial}
    & \suiteheader{\compicon}{Composite}{Reflect-10}
    & \suiteheader{\cubeicon}{Object Binding}{Reflect-Object}
    & \textbf{Mean} \\
    \midrule
    \modelcell{\qwenicon}{\modelname{QwenPI-Flow}} & 0.0 & 7.2 & 1.8 & 0.0 & 2.25 \\
    \modelcell{\llamaicon}{\modelname{OpenVLA-OFT}} & 4.6 & 7.2 & 0.0 & 1.2 & 3.25 \\
    \modelcell{\llamaicon}{\modelname{OpenVLA}} & 0.0 & 0.4 & 0.0 & 0.0 & 0.10 \\
    \bottomrule
  \end{tabular}
  }%
  \caption{Diagnostic success rates for the challenging episodes sampled from \benchname{LIBERO-PRO} and retained in \benchname{LIBERO-Reflect}. Values are percentages and summarize the challenging side used by the benchmark.}
  \label{tab:libero-pro-negative-success}
\end{table*}

\section{Ablation Details and Exact Values}
\label{app:ablation-source-data}

This appendix provides the exact values for the ablation figures and explains how to read the trends. The ablations are not additional methods; they test whether \ourmethod depends on broad head averaging or manually selected layer ranges. The main configuration deliberately uses all layers with \topkname{1} head selection, because this setting preserves the strongest and most stable architecture-aware signal while adding the least aggregation overhead.

\paragraph{Text versus visual heads.}
Table~\ref{tab:text-visual-head-source} reports the exact values for the text-head versus visual-head comparison in the main paper. We keep only the text-head \topkname{1} counterpart and use the same orientation terminology as the main method: \maed{} for \openvla and \maec{} for \qwenpiflow. The comparison shows why text-side entropy is insufficient: it can be high on subsets where language-side uncertainty is predictive, but the large visual advantages on \openvla Object Binding and \qwenpiflow Goal Semantics/\subsetname{Reflect-10} show that it misses visually grounded action failures.

\begin{table*}[t!]
  \centering
  \small
  \setlength{\tabcolsep}{2pt}
  \renewcommand{\arraystretch}{1.18}
  \begin{tabular}{>{\raggedright\arraybackslash}p{0.18\textwidth}
                  >{\raggedright\arraybackslash}p{0.15\textwidth}
                  >{\centering\arraybackslash}p{0.10\textwidth}
                  >{\centering\arraybackslash}p{0.10\textwidth}
                  >{\centering\arraybackslash}p{0.10\textwidth}
                  >{\centering\arraybackslash}p{0.10\textwidth}
                  >{\centering\arraybackslash}p{0.09\textwidth}}
    \toprule
    \textbf{Model} & \textbf{Subset} & \textbf{Text Head} & \makecell[c]{\textbf{Text}\\\textbf{\metricname{AUROC}}} & \textbf{Visual Head} & \makecell[c]{\textbf{Visual}\\\textbf{\metricname{AUROC}}} & \(\Delta\)\textbf{\metricname{AUROC}} \\
    \midrule
    \multicolumn{7}{c}{\textbf{\ding{182} Latent-Readout VLA}} \\
    \midrule
    \multirow{4}{*}{\modelcell{\llamaicon}{\modelname{OpenVLA}}}
    & \subsetname{Reflect-Goal} & \maed{}~\topkname{1} & 59.26 & \maemethodcell{\maed{}~\topkname{1}} & \shadebf{63.94} & \textcolor{maePurple!75!black}{\textbf{+4.68}} \\
    & \subsetname{Reflect-Object} & \maed{}~\topkname{1} & 69.41 & \maemethodcell{\maed{}~\topkname{1}} & \shadebf{90.97} & \textcolor{maePurple!75!black}{\textbf{+21.56}} \\
    & \subsetname{Reflect-Spatial} & \maed{}~\topkname{1} & 62.31 & \maemethodcell{\maed{}~\topkname{1}} & \shadebf{66.86} & \textcolor{maePurple!75!black}{\textbf{+4.55}} \\
    & \subsetname{Reflect-10} & \maed{}~\topkname{1} & 49.43 & \maemethodcell{\maed{}~\topkname{1}} & \shadebf{54.74} & \textcolor{maePurple!75!black}{\textbf{+5.31}} \\
    \midrule
    \multicolumn{7}{c}{\textbf{\ding{183} Latent-Refinement VLA}} \\
    \midrule
    \multirow{4}{*}{\modelcell{\qwenicon}{\modelname{QwenPI-Flow}}}
    & \subsetname{Reflect-Goal} & \maec{}~\topkname{1} & 55.46 & \maemethodcell{\maec{}~\topkname{1}} & \shadebf{80.57} & \textcolor{maePurple!75!black}{\textbf{+25.11}} \\
    & \subsetname{Reflect-Object} & \maec{}~\topkname{1} & 66.00 & \maemethodcell{\maec{}~\topkname{1}} & \shadebf{75.94} & \textcolor{maePurple!75!black}{\textbf{+9.94}} \\
    & \subsetname{Reflect-Spatial} & \maec{}~\topkname{1} & 84.25 & \maemethodcell{\maec{}~\topkname{1}} & \shadebf{84.80} & \textcolor{maePurple!75!black}{\textbf{+0.55}} \\
    & \subsetname{Reflect-10} & \maec{}~\topkname{1} & 56.91 & \maemethodcell{\maec{}~\topkname{1}} & \shadebf{79.52} & \textcolor{maePurple!75!black}{\textbf{+22.61}} \\
    \bottomrule
  \end{tabular}
  \caption{Exact \metricname{AUROC} values for the text-head versus visual-head comparison in the main paper. Purple cells mark the visual-head signal used by \ourmethod, and purple deltas report visual-head \ourmethod{} minus text-head \ourmethod{} under the same \topkname{1} setting. Subset names use the \benchname{LIBERO-Reflect} split identifiers.}
  \label{tab:text-visual-head-source}
\end{table*}

\paragraph{Head selection.}
Table~\ref{tab:vla-aggregation-results} reports the full \topkname{m} head-selection sweep used in the main paper. Across all three models, \topkname{1} is the most reliable default when considering both \metricname{AUROC} and \metricname{AUPR}. Increasing \(m\) admits more heads, but the additional heads are not guaranteed to be action-relevant; in practice they often dilute the oriented entropy signal. The half-head setting remains competitive in some subsets, yet it is less consistent and costs more aggregation, so we use \topkname{1} in the main protocol.

\begin{table*}[t!]
  \centering
  \small
  \setlength{\tabcolsep}{0pt}
  \renewcommand{\arraystretch}{1.18}
  \resizebox{\textwidth}{!}{%
  \begin{tabular}{>{\centering\arraybackslash}p{28mm}
                  >{\centering\arraybackslash}p{22mm}
                  *{12}{>{\centering\arraybackslash}p{9.4mm}}}
    \toprule
    \multirow{2}{*}{Model} & \multirow{2}{22mm}{\makecell[c]{\topkname{m}\\Setting}}
    & \multicolumn{3}{c}{\suiteheader{\goalicon}{Goal Semantics}{Reflect-Goal}}
    & \multicolumn{3}{c}{\suiteheader{\cubeicon}{Object Binding}{Reflect-Object}}
    & \multicolumn{3}{c}{\suiteheader{\routeicon}{Spatial Grounding}{Reflect-Spatial}}
    & \multicolumn{3}{c}{\suiteheader{\compicon}{Composite}{Reflect-10}} \\
    \cmidrule(lr){3-5} \cmidrule(lr){6-8} \cmidrule(lr){9-11} \cmidrule(lr){12-14}
    & & {\footnotesize\metricname{AUROC}} & {\footnotesize\metricname{AUPR}} & \makecell[c]{\footnotesize\metricname{FPR}\\[-0.15em]\footnotesize @95}
      & {\footnotesize\metricname{AUROC}} & {\footnotesize\metricname{AUPR}} & \makecell[c]{\footnotesize\metricname{FPR}\\[-0.15em]\footnotesize @95}
      & {\footnotesize\metricname{AUROC}} & {\footnotesize\metricname{AUPR}} & \makecell[c]{\footnotesize\metricname{FPR}\\[-0.15em]\footnotesize @95}
      & {\footnotesize\metricname{AUROC}} & {\footnotesize\metricname{AUPR}} & \makecell[c]{\footnotesize\metricname{FPR}\\[-0.15em]\footnotesize @95} \\
    \midrule
    \multicolumn{14}{c}{\textbf{\ding{182} Latent-Readout VLAs}} \\
    \midrule
    \multirow{4}{28mm}{\modelcell{\llamaicon}{\modelname{OpenVLA}}}
    & \maemethodcell{\maed{}~\topkname{1}} & \textbf{63.94} & \textbf{43.23} & \textbf{61.92} & \textbf{90.97} & \textbf{75.88} & \textbf{28.30} & \textbf{66.86} & \textbf{50.74} & \textbf{75.79} & \textbf{54.74} & \textbf{30.45} & \textbf{86.46} \\
    \maeboxrule
    & \maed{}~\topkname{2} & 62.41 & 42.33 & 68.21 & 89.14 & 72.76 & 30.05 & 64.72 & 52.29 & 77.13 & 54.01 & 26.97 & 85.91 \\
    & \maed{}~\topkname{4} & 60.36 & 41.01 & 70.03 & 87.13 & 69.26 & 34.18 & 64.25 & 52.14 & 78.80 & 53.20 & 26.55 & 86.88 \\
    \maeboxrule
    & \maemethodcell{\maed{}~\topkname{16}} & \underline{59.56} & \underline{40.67} & \underline{71.52} & \underline{79.64} & \underline{58.98} & \underline{48.81} & \underline{63.99} & \underline{49.67} & \underline{84.81} & \underline{53.21} & \underline{26.56} & \underline{87.43} \\
    \arrayrulecolor{maePurple}\midrule\arrayrulecolor{black}
    \multirow{4}{28mm}{\modelcell{\llamaicon}{\modelname{OpenVLA-OFT}}}
    & \maemethodcell{\maed{}~\topkname{1}} & \textbf{97.34} & \textbf{96.14} & \textbf{7.14} & \textbf{80.56} & \textbf{81.10} & \textbf{63.71} & \textbf{92.63} & \textbf{93.15} & \textbf{39.31} & \textbf{78.57} & \textbf{64.08} & \textbf{45.06} \\
    \maeboxrule
    & \maed{}~\topkname{2} & 96.86 & 95.26 & 7.76 & 76.74 & 77.29 & 65.32 & 93.10 & 93.59 & 38.10 & 77.86 & 62.92 & 43.73 \\
    & \maed{}~\topkname{4} & 94.63 & 94.38 & 8.37 & 78.42 & 79.10 & 63.71 & 90.25 & 93.01 & 35.48 & 77.79 & 62.85 & 46.39 \\
    \maeboxrule
    & \maemethodcell{\maed{}~\topkname{16}} & \underline{91.84} & \underline{94.50} & \underline{7.94} & \underline{75.96} & \underline{82.86} & \underline{64.03} & \underline{91.71} & \underline{89.85} & \underline{42.74} & \underline{77.13} & \underline{62.10} & \underline{45.06} \\
    \arrayrulecolor{maePurple}\midrule\arrayrulecolor{black}
    \multicolumn{14}{c}{\textbf{\ding{183} Latent-Refinement VLAs}} \\
    \midrule
    \multirow{4}{28mm}{\modelcell{\qwenicon}{\modelname{QwenPI-Flow}}}
    & \maemethodcell{\maec{}~\topkname{1}} & \textbf{80.57} & \textbf{80.01} & \textbf{60.12} & \textbf{75.94} & \textbf{76.48} & \textbf{81.18} & \textbf{84.80} & \textbf{85.46} & \textbf{68.53} & \textbf{79.52} & \textbf{76.24} & \textbf{55.60} \\
    \maeboxrule
    & \maec{}~\topkname{2} & 78.75 & 84.28 & 55.45 & 73.01 & 72.27 & 82.35 & 85.73 & 82.67 & 53.00 & 77.75 & 72.72 & 56.18 \\
    & \maec{}~\topkname{4} & 74.16 & 81.17 & 64.79 & 63.42 & 59.80 & 86.27 & 79.63 & 79.03 & 68.12 & 77.47 & 70.39 & 59.27 \\
    \maeboxrule
    & \maemethodcell{\maec{}~\topkname{20}} & \underline{62.63} & \underline{59.16} & \underline{80.93} & \underline{65.92} & \underline{63.43} & \underline{81.59} & \underline{66.89} & \underline{64.70} & \underline{73.50} & \underline{68.91} & \underline{61.92} & \underline{74.52} \\
    \arrayrulecolor{maePurple}\bottomrule\arrayrulecolor{black}
  \end{tabular}
  }%
  \caption{\topkname{m} head-selection ablation values for the main paper. Bold values mark the \topkname{1} setting used in the main results, while underlined values mark the half-head comparison: \topkname{16} for 32-head OpenVLA-family models and \topkname{20} for the 40-head QwenPI-Flow model. \topkname{1} gives the most stable ranking quality while avoiding noisy aggregation over many heads.}
  \label{tab:vla-aggregation-results}
\end{table*}

\paragraph{Layer bands.}
Table~\ref{tab:openvla-layer-source} reports the exact values for the layer-band ablation in the main paper. Restricting the score to shallow, middle, or deep layers can preserve parts of the signal, especially on subsets where object grounding is already strongly localized. However, no restricted band dominates across task families. The all-layer score gives the best overall \metricname{AUROC} and avoids tuning a layer range per architecture or per task, which is important for a self-evaluation method intended to transfer across VLA backbones.

\begin{table*}[t!]
  \centering
  \small
  \setlength{\tabcolsep}{2pt}
  \renewcommand{\arraystretch}{1.18}
  \begin{tabular}{lccccc}
    \toprule
    \textbf{Layer Band}
    & \makecell[c]{\textbf{Overall}\\[-0.15em]\small All subsets}
    & \suiteheader{\compicon}{Composite}{Reflect-10}
    & \suiteheader{\goalicon}{Goal Semantics}{Reflect-Goal}
    & \suiteheader{\cubeicon}{Object Binding}{Reflect-Object}
    & \suiteheader{\routeicon}{Spatial Grounding}{Reflect-Spatial} \\
    \arrayrulecolor{maePurple}\midrule\arrayrulecolor{black}
    \maemethodcell{\textbf{All layers}} & \shadebf{67.88} & \shadebf{54.74} & \shadebf{63.94} & \shadebf{90.97} & \shadebf{66.86} \\
    \arrayrulecolor{maePurple}\midrule\arrayrulecolor{black}
    Shallow & 65.82 & 51.49 & 57.00 & 90.43 & 64.97 \\
    Middle & 66.88 & 54.25 & 61.93 & 86.59 & 65.28 \\
    Deep & 62.91 & 47.71 & 61.49 & 83.50 & 65.55 \\
    \bottomrule
  \end{tabular}
  \caption{OpenVLA layer-band ablation values for the main paper. Values are \maed{} with \topkname{1} \metricname{AUROC} percentages. The all-layer setting is the default configuration because it is strongest overall and avoids task-specific layer tuning.}
  \label{tab:openvla-layer-source}
\end{table*}

\section{Evaluation Protocol}
\label{app:evaluation-settings}

All self-evaluation methods are compared under the same episode-level protocol. A policy first executes a \benchname{LIBERO-Reflect} episode, and the simulator success flag defines the binary label. The self-evaluation method then assigns a scalar reliability score to that episode without using the label. We evaluate whether the score ranks successful executions above failed executions using \metricname{AUROC}, \metricname{AUPR}, and \metricname{FPR@95}. For \ourmethod, attention maps are taken from the same policy forward passes that generate the robot actions; no auxiliary model, additional rollout, or external verifier is required. Table~\ref{tab:evaluation-settings} summarizes the model-specific attention dimensions and the score orientation used for each policy.

\begin{table*}[t!]
  \centering
  \small
  \setlength{\tabcolsep}{2pt}
  \renewcommand{\arraystretch}{1.18}
  \begin{tabular}{p{0.18\textwidth}p{0.17\textwidth}p{0.11\textwidth}p{0.14\textwidth}p{0.18\textwidth}p{0.16\textwidth}}
    \toprule
    \textbf{Policy} & \textbf{Action-generation family} & \textbf{Attention depth} & \textbf{Main score} & \textbf{Head-selection settings} & \textbf{Evaluation role} \\
    \midrule
    \modelcell{\llamaicon}{\modelname{OpenVLA}} & Latent-Readout VLAs & 32 layers / 32 heads & \maemethodcell{\maed{}~\topkname{1}} & \topkname{1}, \topkname{2}, \topkname{4}, \topkname{16} & Autoregressive discrete readout; supports token-statistic baselines \\
    \midrule
    \modelcell{\llamaicon}{\modelname{OpenVLA-OFT}} & Latent-Readout VLAs & 32 layers / 32 heads & \maemethodcell{\maed{}~\topkname{1}} & \topkname{1}, \topkname{2}, \topkname{4}, \topkname{16} & Same VLA family with continuous readout \\
    \midrule
    \modelcell{\qwenicon}{\modelname{QwenPI-Flow}} & Latent-Refinement VLAs & 36 layers / 40 heads & \maemethodcell{\maec{}~\topkname{1}} & \topkname{1}, \topkname{2}, \topkname{4}, \topkname{20} & Cross-family flow-style refinement policy \\
    \bottomrule
  \end{tabular}
  \caption{Evaluation settings for \ourmethod{} across the three VLA policies. The main score uses all layers with \topkname{1} head selection. Larger \topkname{m} settings are reported only for the head-selection ablation; the half-head setting is \topkname{16} for 32-head OpenVLA-family policies and \topkname{20} for the 40-head QwenPI-Flow policy.}
  \label{tab:evaluation-settings}
\end{table*}

The transformer layer index is denoted by \(\ell\), while \(k\) denotes the internal action-generation step.
Table~\ref{tab:internal-step-correspondence} summarizes the model-specific correspondence between \(k\) and the final internal step \(K\) used by \ourmethod{}.

\begin{table*}[t!]
  \centering
  \small
  \setlength{\tabcolsep}{3pt}
  \renewcommand{\arraystretch}{1.18}
  \begin{tabular}{p{0.20\textwidth}p{0.25\textwidth}p{0.26\textwidth}p{0.19\textwidth}}
    \toprule
    \textbf{Policy} & \textbf{Internal generation step \(k\)} & \textbf{Final internal step \(K\) used by \ourmethod{}} & \textbf{\(K\) in our implementation} \\
    \midrule
    \modelcell{\llamaicon}{\modelname{OpenVLA}} & One autoregressive action-token generation step & The last action-token generation step & Number of generated action tokens \\
    \midrule
    \modelcell{\llamaicon}{\modelname{OpenVLA-OFT}} & One continuous-readout step & The continuous-readout step before the action head produces the action chunk & \(1\) \\
    \midrule
    \modelcell{\qwenicon}{\modelname{QwenPI-Flow}} & One flow-style refinement step & The last refinement step before the action trajectory is returned & Number of inference refinement steps \\
    \bottomrule
  \end{tabular}
  \caption{Model-specific meaning of the internal generation step \(k\) and the final internal step \(K\) used by \ourmethod{}.}
  \label{tab:internal-step-correspondence}
\end{table*}

\paragraph{Layer-band analysis.}
The main protocol aggregates visual attention entropy over all layers. For the layer-band ablation, we additionally divide the network depth into shallow, middle, and deep regions and recompute the same \maed{} or \maec{} score within each region. This isolates whether the reliability signal is concentrated at a particular depth or benefits from integrating the full action-generation process. The corresponding exact values are reported in \appref{app:ablation-source-data}.

\paragraph{Random baseline.}
The random baseline assigns an independent uniform score to every episode under the same labels over successful and failed episodes. It is used only as a ranking lower bound and is not tuned per subset or per model.

\section{Baseline Protocols and Applicability}
\label{app:baseline-applicability}

This appendix defines the baselines used in the main results table and clarifies their applicability to heterogeneous VLA action heads. We only discuss baselines that are part of the reported protocol. The main compatibility issue is whether a policy exposes autoregressive discrete action-token logits or sampled action-token sequences. Token-statistic baselines are meaningful for discrete \modelname{OpenVLA}, but they are not defined for \modelname{OpenVLA-OFT}'s continuous action head or \modelname{QwenPI-Flow}'s flow-matching action head without adding a separate likelihood model over executed actions. Token-level Self-Consistency has the same interface requirement because it measures sampled action-token agreement. The main results table omits technically not-applicable rows for continuous-action and flow-action policies, while this appendix documents the interface mismatch behind those omissions.

\paragraph{Random.}
Random assigns an i.i.d. uniform score to each episode and serves as a lower-bound ranking baseline under the same labels over successful and failed episodes.

\paragraph{Verbal Confidence.}
Verbal Confidence adapts p(True)-style verbal self-checking to VLA evaluation. Since most VLA policies do not naturally output a calibrated verbal probability during action generation, we use an external multimodal evaluator as a proxy. For each episode, we query gpt-4.1 with the task instruction and a stitched contact-sheet image of sampled observations. The model returns a scalar confidence in \([0,1]\), which is used directly as the episode-level reliability score. The evaluator is not given the ground-truth success label, simulator success state, object poses, perturbation metadata, robot state trajectories, generated action vectors, or oracle information.

\begin{figure*}[t]
\centering
\begin{tcolorbox}[
  enhanced,
  colback=maePurple!4,
  colframe=maePurple!70!black,
  coltitle=white,
  fonttitle=\bfseries,
  title={Verbal Confidence Prompt},
  boxrule=0.6pt,
  arc=2pt,
  boxsep=0.7em,
  left=0.7em,
  right=0.7em,
  top=0.7em,
  bottom=0.7em,
  width=\textwidth
]
\small
\textbf{System context:} You are given a robot manipulation episode.\\[0.25em]
\textbf{Image input:} The attached image is a stitched contact sheet of observation frames sampled from the episode. Frames are ordered chronologically from left to right and top to bottom. Each frame is labeled with its timestep.\\[0.35em]
\textbf{Task instruction:}\\
\{TASK\_INSTRUCTION\}\\[0.35em]
\textbf{Evaluation request:} Based on the task instruction and the attached contact-sheet image, estimate whether the robot successfully completed the task.\\[0.25em]
\textbf{Output format:} Return only one number between 0 and 1: 0 means definitely failed, 0.5 means uncertain, and 1 means definitely succeeded. Return only the number and nothing else.
\end{tcolorbox}
\caption{Prompt used for the Verbal Confidence baseline. The stitched contact sheet is provided as an image input in the same multimodal API request.}
\label{fig:verbal-confidence-prompt}
\end{figure*}

\paragraph{Self-Consistency.}
Self-Consistency is implemented as token-level sampled action-token agreement for discrete \modelname{OpenVLA}. At each execution step, we sample \(n\) action-token sequences from the same observation and task condition. For each action-token dimension, the score is the frequency of the most common sampled token divided by \(n\); the step score averages this value over action-token dimensions, and the episode score averages over execution steps. The first sampled action is executed, while the additional samples are used only to compute the self-evaluation score.

This baseline requires discrete sampled action tokens. \modelname{OpenVLA-OFT} predicts executed actions through a continuous action head, and \modelname{QwenPI-Flow} produces continuous action trajectories through flow-matching refinement. Their action heads do not expose sampled categorical action-token sequences for the executed action, so token-level Self-Consistency is not defined for these two policies in the reported protocol.

\paragraph{Maximum Softmax Probability.}
Maximum Softmax Probability measures the sharpness of autoregressive action-token predictions. For a generated action-token sequence of length \(T\), with categorical distribution \(y_t\) over the vocabulary at generation step \(t\), the episode score is
\[
  \frac{1}{T}\sum_{t=1}^{T}\max_i y_{t,i}.
\]
This requires a vocabulary-level distribution for each generated action token, so it applies to discrete OpenVLA but not to continuous OFT or flow-matching PI action heads.

\paragraph{Perplexity.}
Perplexity-style confidence uses the negative log confidence of the selected action-token distribution:
\[
  \frac{1}{T}\sum_{t=1}^{T}-\log\left(\max_i y_{t,i}\right).
\]
We invert the direction when necessary so that larger reported scores indicate higher estimated reliability. As with Maximum Softmax Probability, this score is only defined when action generation exposes token-level categorical probabilities.

\paragraph{Entropy.}
Entropy measures the uncertainty of the full output distribution:
\[
  \frac{1}{T}\sum_{t=1}^{T}\sum_i -y_{t,i}\log y_{t,i}.
\]
Lower entropy indicates a sharper token distribution, so the reliability score uses the sign convention that larger is better. This is a token-distribution baseline and is not comparable for continuous action regression or flow integration without an additional probabilistic action model.

\paragraph{Length-normalized Entropy.}
Length-normalized entropy generates \(n\) candidate outputs \(\mathcal{Y}=\{Y_1,\dots,Y_n\}\), then averages token entropy across the sampled outputs:
\[
  \frac{1}{n}\sum_{Y\in\mathcal{Y}}
    \frac{1}{T_Y}\sum_{t=1}^{T_Y}
      \sum_i -y_{t,i}\log y_{t,i}.
\]
We set \(n=5\).

\paragraph{OpenVLA output-statistic values.} We report reliability-oriented scores: Perplexity is represented by inverse PPL, Entropy by inverse token entropy, and Length-normalized Entropy by inverse length-normalized entropy, so larger values always indicate higher estimated reliability. These baselines can be computed for \modelname{OpenVLA} because it generates discrete action tokens with vocabulary-level probabilities; applying them to \modelname{OpenVLA-OFT} or \modelname{QwenPI-Flow} would require adding a separate likelihood model over continuous executed actions.

\paragraph{Applicability to heterogeneous VLAs.}
Discrete OpenVLA exposes autoregressive action-token distributions, so token-probability baselines such as Maximum Softmax Probability, Perplexity, Entropy, and Length-normalized Entropy are conceptually defined for that model family. OpenVLA-OFT changes the action interface: it keeps the OpenVLA backbone but predicts continuous actions through a regression head, so the placeholder action slots are not generated action tokens and their logits do not define the executed action. QwenPI-Flow uses a flow-matching action head that maps visual-language hidden states and noise through iterative velocity prediction; its output is a continuous action trajectory rather than a categorical token sequence. Applying token-probability baselines to these models would evaluate a different random variable from the executed action. By contrast, \ourmethod reads internal attention entropy during the same policy forward process and then uses architecture-aware aggregation, \maed for Latent-Readout VLAs and \maec for Latent-Refinement VLAs, which is why it remains comparable across the heterogeneous action heads in the main results table.

\section{Model Input Templates}
\label{app:chat-templates}

Table~\ref{tab:model-chat-templates} reports the policy input formats used to condition the three evaluated VLA backbones. The OpenVLA-family policies use the same action-query prompt form with lower-cased LIBERO instructions, while \modelname{OpenVLA-OFT} additionally uses proprioceptive State Input for continuous readout. \modelname{QwenPI-Flow} uses a Qwen3-VL-style multimodal message and appends the grounding text expected by the StarVLA policy. We place these templates after the experimental tables because they are protocol details rather than additional results.

\begin{table*}[t!]
  \centering
  \small
  \setlength{\tabcolsep}{1pt}
  \renewcommand{\arraystretch}{1.18}
  \begin{tabular}{>{\raggedright\arraybackslash}p{0.20\textwidth}
                  >{\raggedright\arraybackslash}p{0.22\textwidth}
                  >{\raggedright\arraybackslash}p{0.24\textwidth}
                  >{\raggedright\arraybackslash}p{0.28\textwidth}}
    \toprule
    \textbf{Policy} & \textbf{Input convention} & \textbf{Episode payload} & \textbf{Rendered prompt form} \\
    \midrule
    \modelcell{\llamaicon}{\modelname{OpenVLA}}
    & Pure action-prompt format used by the OpenVLA policy.
    & One RGB observation image and the lower-cased LIBERO task instruction.
    & \promptcode{In: What action should the robot take to \{instruction.lower()\}?}\newline
      \promptcode{Out:} \\
    \midrule
    \modelcell{\llamaicon}{\modelname{OpenVLA-OFT}}
    & OpenVLA-family action prompt with the OFT continuous readout.
    & Primary image, wrist image, proprioceptive State Input, and the lower-cased task label.
    & \promptcode{In: What action should the robot take to \{task\_label.lower()\}?}\newline
      \promptcode{Out:} \\
    \midrule
    \modelcell{\qwenicon}{\modelname{QwenPI-Flow}}
    & Qwen3-VL multimodal message format followed by the StarVLA grounding request.
    & One or more image placeholders followed by the LIBERO instruction and object-localization request.
    & \promptcode{<|im\_start|>user}\newline
      \promptcode{<|vision\_start|><|image\_pad|><\newline|vision\_end|>}\newline
      \promptcode{Your task is \{instruction\}. To identify the key objects for your task. Locate their bounding boxes in [x1,y1,x2,y2] format.}\newline
      \promptcode{<|im\_end|>}\newline
      \promptcode{<|im\_start|>assistant} \\
    \bottomrule
  \end{tabular}
  \caption{Model input templates used for policy conditioning. The table records the prompt forms and non-text policy inputs needed to reproduce the action-generation inputs; benchmark construction and experimental results are reported in \apprefs{app:libero-reflect-construction}{app:ablation-source-data}.}
  \label{tab:model-chat-templates}
\end{table*}

\begin{table*}[t]
  \centering
  \tiny
  \setlength{\tabcolsep}{1.7pt}
  \renewcommand{\arraystretch}{1.12}
  \resizebox{\textwidth}{!}{%
  \begin{tabular}{ll*{15}{c}}
    \toprule
    \textbf{Category} & \textbf{Method}
    & \multicolumn{3}{c}{\textbf{\subsetname{Reflect-Goal}}}
    & \multicolumn{3}{c}{\textbf{\subsetname{Reflect-Object}}}
    & \multicolumn{3}{c}{\textbf{\subsetname{Reflect-Spatial}}}
    & \multicolumn{3}{c}{\textbf{\subsetname{Reflect-10}}}
    & \multicolumn{3}{c}{\textbf{Overall ($N=1{,}116$)}} \\
    \cmidrule(lr){3-5}\cmidrule(lr){6-8}\cmidrule(lr){9-11}\cmidrule(lr){12-14}\cmidrule(lr){15-17}
    & & \metricname{AUROC}$\uparrow$ & \metricname{AUPR}$\uparrow$ & \metricname{FPR@95}$\downarrow$
      & \metricname{AUROC}$\uparrow$ & \metricname{AUPR}$\uparrow$ & \metricname{FPR@95}$\downarrow$
      & \metricname{AUROC}$\uparrow$ & \metricname{AUPR}$\uparrow$ & \metricname{FPR@95}$\downarrow$
      & \metricname{AUROC}$\uparrow$ & \metricname{AUPR}$\uparrow$ & \metricname{FPR@95}$\downarrow$
      & \metricname{AUROC}$\uparrow$ & \metricname{AUPR}$\uparrow$ & \metricname{FPR@95}$\downarrow$ \\
    \midrule
    \multirow{3}{*}{Black-box}
      & Random & 49.82 & 50.11 & 95.12 & 50.15 & 49.85 & 95.30 & 49.90 & 50.05 & 94.90 & 50.20 & 49.90 & 94.80 & 50.02 & 49.98 & 95.02 \\
      & Verbal Conf. & 50.45 & 50.62 & 93.20 & 50.10 & 50.30 & 93.80 & 51.10 & 51.40 & 92.10 & 52.80 & 52.40 & 92.50 & 51.11 & 51.18 & 92.90 \\
      & Self-Cons. & 56.12 & 55.80 & 85.30 & 57.80 & 57.20 & 83.10 & 57.30 & 56.90 & 88.20 & 53.20 & 53.00 & 91.20 & 56.11 & 55.72 & 86.95 \\
    \midrule
    \multirow{4}{*}{White-box}
      & MaxProb & 54.30 & 54.10 & 89.10 & 55.20 & 54.90 & 87.30 & 55.10 & 54.70 & 90.40 & 52.90 & 52.60 & 92.00 & 54.38 & 54.08 & 89.70 \\
      & Perplexity & 54.80 & 54.55 & 88.40 & 55.60 & 55.20 & 86.80 & 55.50 & 55.10 & 89.80 & 53.05 & 52.80 & 91.60 & 54.74 & 54.41 & 89.15 \\
      & Entropy & 55.70 & 55.40 & 87.20 & 56.80 & 56.40 & 85.30 & 56.70 & 56.30 & 88.70 & 53.40 & 53.10 & 90.90 & 55.65 & 55.30 & 88.04 \\
      & Length Ent. & 56.05 & 55.80 & 86.80 & 57.40 & 57.00 & 84.70 & 57.10 & 56.80 & 88.20 & 53.60 & 53.30 & 90.50 & 56.04 & 55.72 & 87.55 \\
    \arrayrulecolor{maePurple}\midrule\arrayrulecolor{black}
    \multirow{2}{*}{\textcolor{maePurple!85!black}{\textbf{Ours}}}
      & \maemethodcell{\maed{}~\topkname{16}} & 58.90 & 58.20 & 72.80 & 78.40 & 76.50 & 50.20 & 63.20 & 62.70 & 85.40 & 52.80 & 52.50 & 88.10 & 63.32 & 62.48 & 74.12 \\
      & \maemethodcell{\maed{}~\topkname{1}} & \textbf{63.15} & \textbf{62.40} & \textbf{63.20} & \textbf{88.75} & \textbf{86.30} & \textbf{30.15} & \textbf{66.15} & \textbf{65.80} & \textbf{76.40} & \textbf{54.20} & \textbf{53.80} & \textbf{87.10} & \textbf{68.06} & \textbf{67.08} & \textbf{64.21} \\
    \arrayrulecolor{maePurple}\bottomrule\arrayrulecolor{black}
  \end{tabular}%
  }
  \caption{Standard-only balanced evaluation for \openvla. Every positive and negative episode comes from the same standard LIBERO environment, and each subset is balanced 1:1 by matching every natural failure with a randomly sampled success from that subset. Values are percentages. Overall reports global ranking over the pooled 558 successes and 558 failures.}
  \label{tab:standard-only-balanced-results}
\end{table*}
\section{Within-Domain Robustness Evaluation}
\label{app:dataset-source}
The benchmark combines standard and challenging episodes to evaluate reliability across a broad range of execution conditions. All metrics use the actual simulator success flag for each rollout, rather than the nominal episode source, so naturally failed standard episodes and occasionally successful challenging episodes remain part of the evaluation. This label definition directly measures execution outcome while preserving the intended mixed-difficulty coverage.

The construction statistics characterize the complementary difficulty regimes. Table~\ref{tab:libero-reflect-composition} shows high standard-side success, with overall rates of 95.8\% for \modelname{QwenPI-Flow}, 96.3\% for \modelname{OpenVLA-OFT}, and 72.1\% for \modelname{OpenVLA}; Table~\ref{tab:libero-pro-negative-success} reports the challenging-side diagnostic success rates. To verify that the reliability signal is not tied to the mixed-difficulty composition, we additionally evaluate a balanced standard-only protocol below.

This standard-only evaluation provides a controlled within-domain test: every positive and negative episode is drawn from the same standard LIBERO distribution, while the 1:1 class balance fixes the positive prevalence for \metricname{AUPR}. Consistent performance under this protocol demonstrates that the method ranks successful and failed executions within a shared environment distribution.

\subsection{Standard-Only Balanced Evaluation}
\label{app:standard-only-balanced}
\begin{figure*}[t!]
  \centering
  \includegraphics[width=\textwidth]{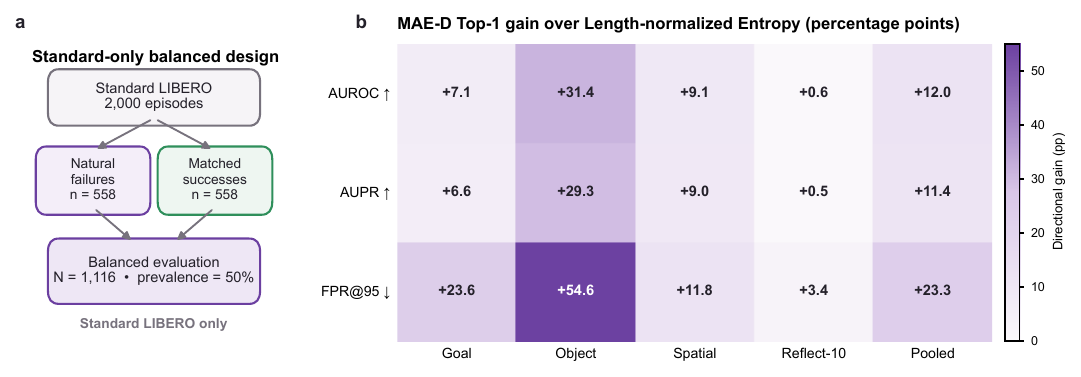}
  \caption{Standard-only balanced evaluation details. \textbf{a}, All 558 natural failures from 2,000 standard LIBERO episodes are paired with 558 subset-matched successes, yielding 1,116 standard-domain episodes at 50\% success prevalence. \textbf{b}, Direction-aware percentage-point gains of \maed{}~\topkname{1} over Length-normalized Entropy across each subset and the pooled set; positive values denote higher \metricname{AUROC}/\metricname{AUPR} or lower \metricname{FPR@95}.}
  \label{fig:standard-only-balanced-details}
\end{figure*}

We construct a complementary evaluation entirely within the standard LIBERO environment. From the 2,000 standard \openvla episodes reported in Table~\ref{tab:libero-reflect-composition}, we retain every natural policy failure: 104 from \subsetname{Reflect-Goal}, 101 from \subsetname{Reflect-Object}, 224 from \subsetname{Reflect-Spatial}, and 129 from \subsetname{Reflect-10}, for 558 failures in total. We then sample an equal number of successful episodes within each subset, without introducing placement swaps or any LIBERO-PRO episode. The resulting set contains 558 successes and 558 failures ($N=1{,}116$), fixing success prevalence at 50\%. Under this balanced protocol, the random expectations are 50\% \metricname{AUROC}, 50\% \metricname{AUPR}, and 95\% \metricname{FPR@95}. Figure~\ref{fig:standard-only-balanced-details} summarizes the sampling construction and the directional gains over the strongest white-box baseline.

Table~\ref{tab:standard-only-balanced-results} reports the full baseline comparison. \maed{}~\topkname{1} is strongest on every subset and pooled metric. Relative to Length-normalized Entropy, the strongest white-box baseline, it improves pooled \metricname{AUROC} from 56.04 to 68.06 and pooled \metricname{AUPR} from 55.72 to 67.08, while reducing pooled \metricname{FPR@95} from 87.55 to 64.21. The largest subset-level separation occurs on \subsetname{Reflect-Object}, where \maed{}~\topkname{1} reaches 88.75 \metricname{AUROC}, 86.30 \metricname{AUPR}, and 30.15 \metricname{FPR@95}. These results establish reliable within-domain ranking among executions drawn from the same standard environment and balanced class prior.

\section{Architecture-Determined Entropy Orientation}
\label{app:orientation}

The opposite entropy orientations used by \maed and \maec are determined by the action-generation interface of the evaluated policy before any episode-level reliability metric is computed.
The distinction follows from the role played by the visual-attention distribution in the transition kernel of the conditional generative Markov chain.

For Latent-Readout VLAs, the internal generation process first builds a latent action representation and then maps the final latent state to an executable action through a readout head.
In this family, the transition kernel must progressively consolidate task-relevant visual evidence into the latent state before the final readout. 
A reliable transition therefore tends to concentrate action-query attention on the relevant object, region, or spatial relation needed for the action. 
Conversely, diffuse visual addressing indicates that the latent action state has not localized the necessary visual evidence, which often corresponds to ambiguous grounding or incorrect object-action binding. 
For this reason, lower visual entropy is assigned higher reliability, yielding the decreasing-entropy orientation \maed.

For Latent-Refinement VLAs, the policy maintains an explicit action or trajectory variable and repeatedly refines it under visual-language conditioning, as in flow- or denoising-style generation.
Here the transition kernel corrects an evolving action trajectory by re-querying the condition across refinement steps.
\ourmethod{} uses the attention at the final internal generation step \(k=K\), before the action trajectory is returned.
Reliable executions retain broader visual addressing at this final refinement step, especially when the correct action depends on object relations, spatial constraints, or multi-step manipulation context.
Overly low visual entropy indicates that the final action trajectory is determined with too narrow a visual context.
Thus, in this family, higher visual entropy at the final refinement step is assigned higher reliability, yielding the increasing-entropy orientation \maec.

This orientation rule is architecture-level rather than data-fitted. 
A model is assigned to Latent-Readout VLAs if its action is produced by reading out a final latent state without iterative refinement of an explicit action variable.
It is assigned to Latent-Refinement VLAs if it maintains an action or trajectory representation that is updated across multiple refinement or denoising steps under repeated conditioning.
This decision can be made from the model's inference computation graph alone and does not require labels over successful and failed episodes, external supervision, or validation-set optimization.
Accordingly, \maed is fixed for Latent-Readout VLAs and \maec is fixed for Latent-Refinement VLAs before evaluating AUROC, AUPR, or FPR@95.

The same principle also separates orientation from head and layer aggregation. 
The Top-1 operation used in the main protocol is a deterministic per-episode reduction over oriented entropy values, not a fixed attention head selected by test labels. 
Likewise, all-layer aggregation is used as the default rule to avoid selecting a task-specific or model-specific layer band. 
The Top-$m$ and layer-band experiments are therefore sensitivity analyses of a frozen scoring rule rather than procedures for choosing the reported orientation.
\section{Cost Analysis Details}
\label{app:cost-analysis-details}

To substantiate the efficiency claims in the main experiments, we profile the wall-clock latency of \ourmethod during evaluation. Table~\ref{tab:cost-analysis-details} reports the average execution time per episode and the extra latency incurred by \ourmethod. The overhead is strictly bounded because \ourmethod reuses the internal attention matrices from the same forward pass and only performs entropy computation and head aggregation.

\begin{table}[t!]
  \centering
  \small
  \setlength{\tabcolsep}{0pt}
  \renewcommand{\arraystretch}{1.18}
  \begin{tabular}{ll}
    \toprule
    \textbf{Component} & \textbf{Detail} \\
    \midrule
    GPU Hardware & NVIDIA H100 \\
    Model Evaluated & \modelcell{\qwenicon}{\modelname{QwenPI-Flow}} \\
    Avg. Rollout Time & $\sim$14.0 s / episode \\
    \arrayrulecolor{maePurple}\midrule\arrayrulecolor{black}
    \maemethodcell{Avg. \ourmethod Extra Time} & \shadebf{0.57 s / episode} \\
    \maemethodcell{\ourmethod Latency Overhead} & \shadebf{4.09\% ($< 0.1\times$)} \\
    \arrayrulecolor{maePurple}\midrule\arrayrulecolor{black}
    Memory Overhead & Negligible (Reuses internal attention) \\
    \bottomrule
  \end{tabular}
  \caption{Empirical cost analysis of \ourmethod{}. The overhead strictly satisfies the $<0.1\times$ boundary highlighted in the Pareto-optimal zone of the main paper.}
  \label{tab:cost-analysis-details}
\end{table}

\FloatBarrier

\end{document}